\documentclass[lettersize,journal]{IEEEtran}
\usepackage{amsmath,amsfonts}
\usepackage{algorithmic}
\usepackage{array}
\usepackage[caption=false,font=normalsize,labelfont=sf,textfont=sf]{subfig}
\usepackage{textcomp}
\usepackage{stfloats}
\usepackage{url}
\usepackage{graphicx}
\usepackage{algorithm}

\usepackage{mathptmx}
\usepackage{cite}
\usepackage{multirow}
\usepackage{tabularx}

\begin{document}
\title{DRHeC: Differentiable Rendering for Hand-Eye Calibration with RGB-Based Gradients
}
\author{Xiaotian Zhang$^{1}$,
    Yusheng Wang$^{2}$,
    Naoya Kagawa$^{3}$,\\
    Noritaka Takamura$^{3}$,
    Keiji Okuhara$^{3}$,
    Hiroyasu Baba$^{3}$,
    Jun Ota$^{2}$
    \thanks{$^{1}$Xiaotian Zhang is with Department of Precision Engineering, School of Engineering, The University of Tokyo, Tokyo, Japan. Email: {\tt\small x.zhang@race.t.u-tokyo.ac.jp}}
    \thanks{$^{2}$Yusheng Wang and Jun Ota are with Research into Artifacts, Center for Engineering (RACE), School of Engineering, The University of Tokyo, Tokyo, Japan}
    \thanks{$^{3}$Naoya Kagawa, Noritaka Takamura, Keiji Okuhara, Hiroyasu Baba are with FA Products Business Unit, DENSO WAVE INCORPORATED, Aichi, Japan}}

\makeatletter
\def\@IEEEpubidpullup{36pt}
\makeatother
\IEEEpubid{\parbox{\textwidth}{\footnotesize
\copyright~2026 IEEE. Personal use of this material is permitted.
Permission from IEEE must be obtained for all other uses, in any current or
future media, including reprinting/republishing this material for advertising
or promotional purposes, creating new collective works, for resale or
redistribution to servers or lists, or reuse of any copyrighted component of
this work in other works.\\
Published in IEEE Transactions on Instrumentation and Measurement,
vol. 75, Art. no. 7505816, pp. 1--16, 2026.
DOI: 10.1109/TIM.2026.3712913
}}
\maketitle

\begin{abstract}
Accurate hand-eye calibration is crucial for precision manipulation.
Traditional methods rely on markers, with their precision dependent on marker accuracy and observability.
In contrast, markerless methods, such as learning-based approaches, use deep neural networks to directly extract keypoints or features from images, enabling the computation of hand-eye transformation with a single image and without the need for physical markers.
However, these methods often face challenges related to dataset scale and data quality.
Recently, differentiable rendering-based methods for hand-eye calibration have leveraged physical models to render binary masks and compare them with observations, enabling hand-eye calibration without fiducial markers in the calibration stage and providing interpretable optimization.
While the state-of-the-art differentiable rendering methods achieve remarkable accuracy, the use of binary masks can result in the loss of internal profile details, reducing precision.
Additionally, these methods are prone to convergence issues during optimization due to the lack of global information in the loss function, which can lead to local minima.

In this study, we propose a novel RGB-based differentiable rendering framework that provides richer geometric and appearance cues by incorporating color and mask geometric features, thereby improving calibration accuracy and optimization stability.
Additionally, we propose a mask-guided image-to-image translation (I2IT) method to ensure explicit preservation of color and geometric consistency throughout the translation.
Our approach is validated through both simulation and real-world experiments, with results demonstrating strong accuracy and robustness and clear improvements over existing differentiable rendering methods.
Our method achieves a grasping success rate of 88.9\% and insertion success rate of 57.4\% on the UR5e real-world experiment, outperforming the state-of-the-art differentiable rendering hand-eye calibration method EasyHeC by 46.3 and 48.1 percentage points, respectively.
\end{abstract}

\begin{IEEEkeywords}
Hand-eye calibration, differentiable rendering, vision-guided robots (VGR), image-to-image translation (I2IT).
\end{IEEEkeywords}

\section{INTRODUCTION}
Robot manipulators are widely used across various fields, and vision-guided robots (VGR) are becoming increasingly popular for pick-and-place tasks due to their superior adaptability and ease of implementation \cite{kofman2005teleoperation, jin2024hand, zhang2020simultaneous, wu2019hand, qiu2020new}. A typical VGR employs cameras as sensors to provide feedback signals, allowing the robot to move precisely to target positions. To accurately determine an object's 3D position and orientation within a robot's workspace, it is crucial to establish the relative poses among several key frames: from the robot base frame to the end-effector frame, from the end-effector frame to the camera frame, and from the camera frame to the object or world frame \cite{sarabandi2022hand, tsai1989new, daniilidis1999hand}. Hand-eye calibration is essential for determining the transformation matrix between the end-effector and camera frames for robotic manipulators \cite{schmidt2008data, fu2020dual, tan2025robust, li2023industrial}. This calibration is essential for vision-guided tasks, as it aligns the camera's coordinate system with the robot's, enabling accurate visual data processing and precise robot control \cite{pedrosa2021general, koide2019general, li2021robot}.

Hand-eye calibration methods can be broadly categorized into marker-based and markerless methods. Traditional marker-based methods rely on specific patterns, such as chessboards, AprilTags, or ChArUco markers, to determine the relative pose between the pattern frame and the camera frame \cite{liu2021precise,ha2022probabilistic,wu2020globally,boby2016single,zhu2024point}.
Their accuracy depends on marker precision, measuring poses, and measurement quality, often requiring extensive design and tuning for specific tasks, which makes them time-consuming and expensive.
\IEEEpubidadjcol
Wang et al. employed dual matrix operators to express the hand-eye calibration equation in the dual quaternion domain, and proposed a new simultaneous method that solves rotation and translation jointly \cite{wang2024dual}.
Jin et al. proposed Single3D, a unified single-marker hand-eye calibration method that avoids intermediate pose estimation and achieves high accuracy and efficiency \cite{jin2024hand}.
In contrast, markerless methods do not require markers, making them easier to execute in real-world robot manipulation tasks.
These methods estimate camera poses by extracting features such as color and texture from the environment, often without prior knowledge of these features.
Although there has been some research into markerless hand-eye calibration, most studies have focused on learning-based approaches that estimate keypoints and use PnP for pose estimation \cite{lu2022pose,lee2020camera}.
These methods face several challenges, including dependence on dataset scale and quality, as well as susceptibility to noise.
Recent studies have introduced differentiable rendering for markerless hand-eye calibration, leveraging physical models to enhance interpretability and generalization with less training data compared to keypoint-based methods \cite{lu2023image, chen2023easyhec, hong2024easyhec++}.
However, differentiable rendering also faces challenges such as vanishing or exploding gradients in the absence of global gradient direction, which can lead to an optimization process that lacks sufficient guidance to accurately converge to the target pose.
This instability can lead to inaccuracies in the hand-eye transformation matrix.
Moreover, relying solely on binary masks in differentiable rendering is inherently problematic, as binary masks contain only silhouette information.
This means that only boundary pixels provide useful gradients, while interior pixels contribute nothing, resulting in an under-constrained optimization.
Pose changes that do not alter the silhouette produce zero gradients, which can lead to vanishing-gradient regions and may also lead to pose ambiguity in some cases where different robot poses produce similar binary masks.

These limitations suggest that binary-mask-based supervision may be insufficient for robust and stable hand-eye calibration in some cases.
To obtain informative and non-degenerate gradients, richer appearance cues such as RGB derivatives and additional global geometric constraints are required to make the optimization both precise and robust.
This observation motivates our framework, which integrates RGB-based differentiable rendering with global geometric features to improve calibration accuracy and optimization stability under binary-mask-based supervision.
This combination is not simply an addition of image features.
By using local RGB appearance cues together with the mask centroid and area terms, it provides more reliable gradient guidance and improves optimization stability.

In this study, we propose a novel framework for hand-eye calibration that leverages differentiable rendering with RGB derivatives and mask geometric features to address these challenges. Given the need for a wide field of view, the simplicity of the experimental setup, and the fact that capturing the entire manipulator in the image improves the robustness of the differentiable rendering process, we primarily adopt the eye-to-hand configuration.
Specifically, our main contributions are:
\begin{itemize}
\item We introduce an integrated differentiable rendering-based calibration framework that combines RGB derivatives with global geometric information, improving optimization accuracy and stability.
\item Within this framework, we design a loss function that explicitly incorporates mask centroids, mask areas, and RGB derivatives to provide reliable gradient guidance.
\item We develop a mask-guided I2IT method that preserves geometric consistency and improves real-to-simulation alignment under mild illumination, color, and material variations.
\item We collect two real robot datasets, each with approximately 1,500 paired images from UR5e and DENSO VS060 for real-to-simulation, to train the real-to-sim translation models.
Real-world experiments on grasping and insertion tasks demonstrate clear improvements over existing binary-mask-based differentiable renderer hand-eye calibration methods.
Comparisons with marker-based baselines are also provided as reference baselines under the tested setup.
\end{itemize}

\section{Related Research}
\subsection{Hand-Eye Calibration}

\begin{figure*}[t]
    \captionsetup[subfloat]{labelformat=empty}
    \centering
    \subfloat[Initial Pose]{\includegraphics[width=0.13\linewidth]{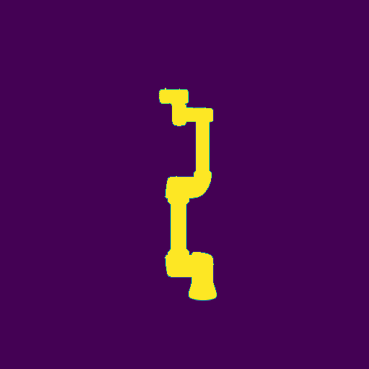}}
    \hspace{0.1mm}
    \subfloat[GT Pose]{\includegraphics[width=0.13\linewidth]{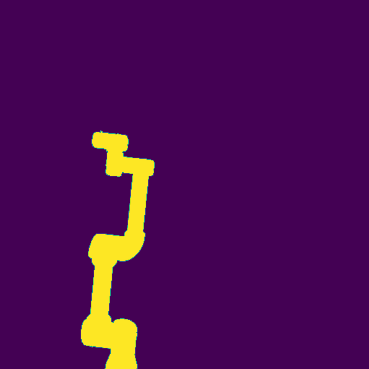}}
    \hspace{0.1mm}
    \subfloat[iter = 100]{\includegraphics[width=0.13\linewidth]{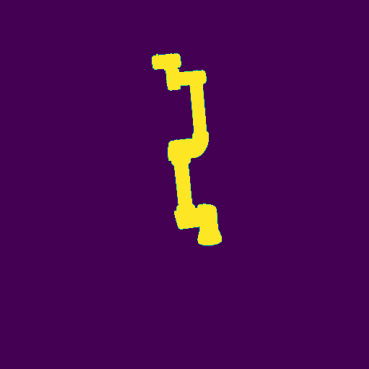}}
    \hspace{0.1mm}
    \subfloat[iter = 200]{\includegraphics[width=0.13\linewidth]{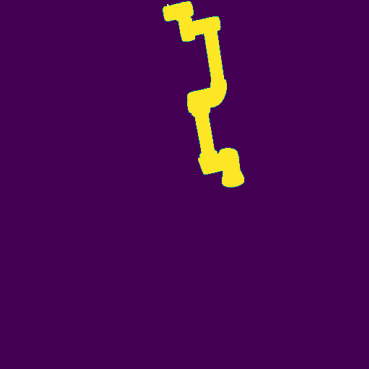}}
    \hspace{0.1mm}
    \subfloat[iter = 300]{\includegraphics[width=0.13\linewidth]{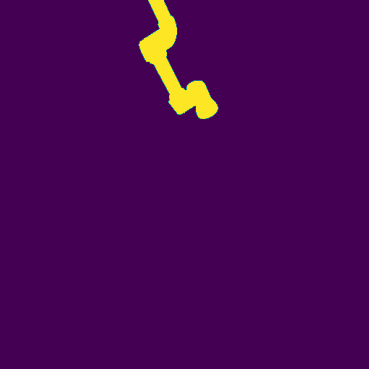}}
    \hspace{0.1mm}
    \subfloat[iter = 400]{\includegraphics[width=0.13\linewidth]{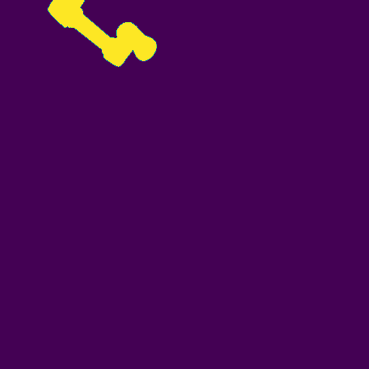}}
    \hspace{0.1mm}
    \subfloat[iter = 500]{\includegraphics[width=0.13\linewidth]{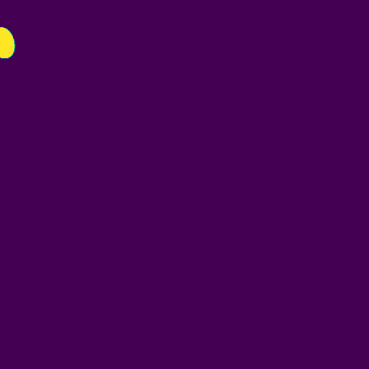}} \\

    \subfloat[Initial Pose]{\includegraphics[width=0.13\linewidth]{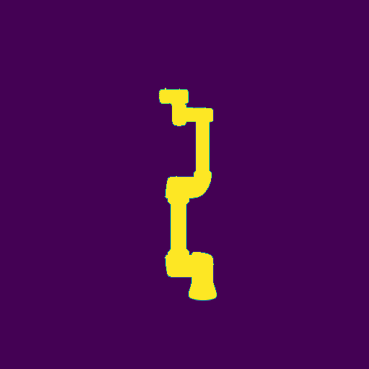}}
    \hspace{0.1mm}
    \subfloat[GT Pose]{\includegraphics[width=0.13\linewidth]{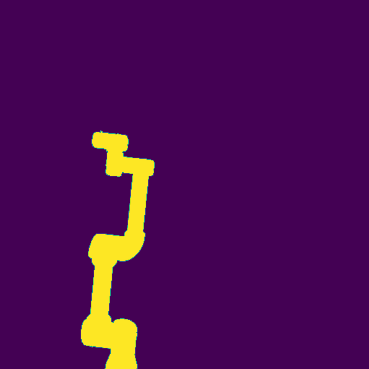}}
    \hspace{0.1mm}
    \subfloat[iter = 100]{\includegraphics[width=0.13\linewidth]{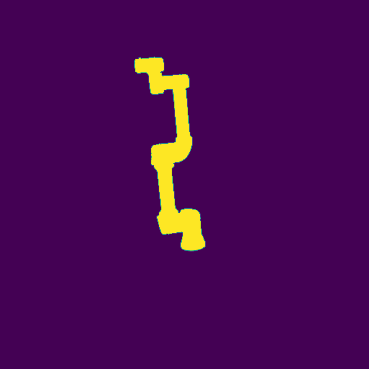}}
    \hspace{0.1mm}
    \subfloat[iter = 200]{\includegraphics[width=0.13\linewidth]{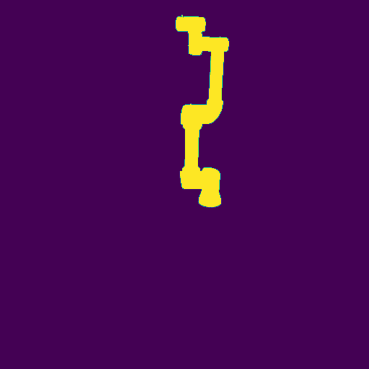}}
    \hspace{0.1mm}
    \subfloat[iter = 300]{\includegraphics[width=0.13\linewidth]{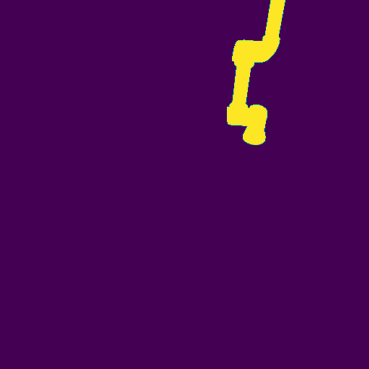}}
    \hspace{0.1mm}
    \subfloat[iter = 400]{\includegraphics[width=0.13\linewidth]{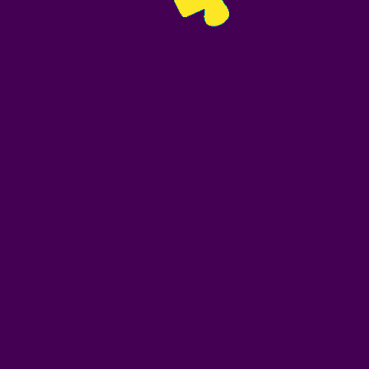}}
    \hspace{0.1mm}
    \subfloat[iter = 500]{\includegraphics[width=0.13\linewidth]{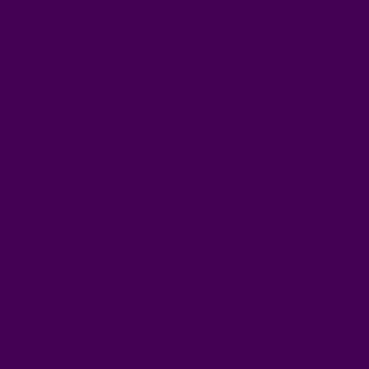}} \\

    \subfloat[Initial Pose]{\includegraphics[width=0.13\linewidth]{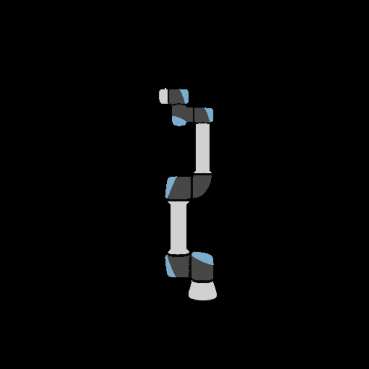}}
    \hspace{0.1mm}
    \subfloat[GT Pose]{\includegraphics[width=0.13\linewidth]{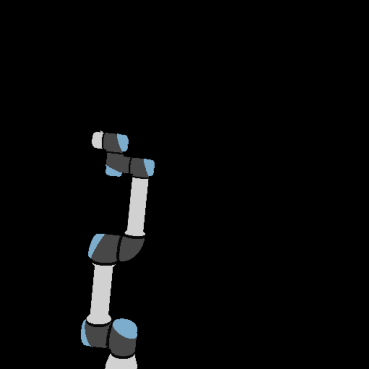}}
    \hspace{0.1mm}
    \subfloat[iter = 100]{\includegraphics[width=0.13\linewidth]{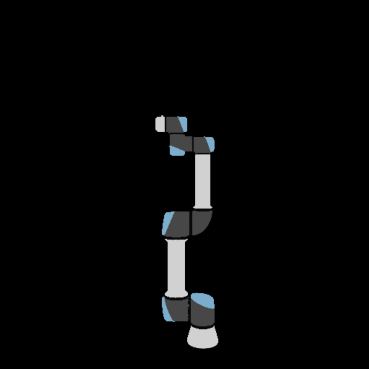}}
    \hspace{0.1mm}
    \subfloat[iter = 200]{\includegraphics[width=0.13\linewidth]{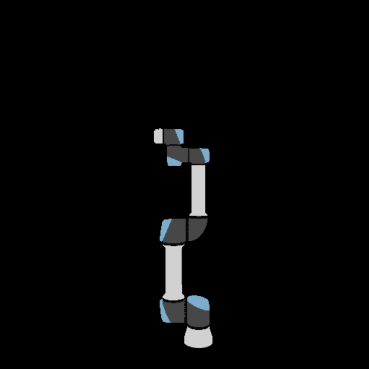}}
    \hspace{0.1mm}
    \subfloat[iter = 300]{\includegraphics[width=0.13\linewidth]{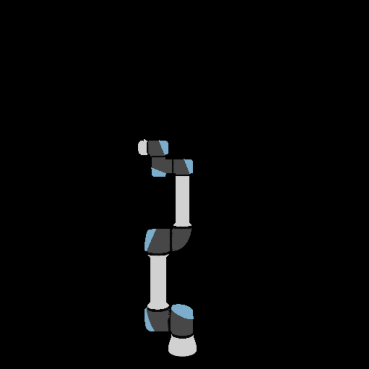}}
    \hspace{0.1mm}
    \subfloat[iter = 400]{\includegraphics[width=0.13\linewidth]{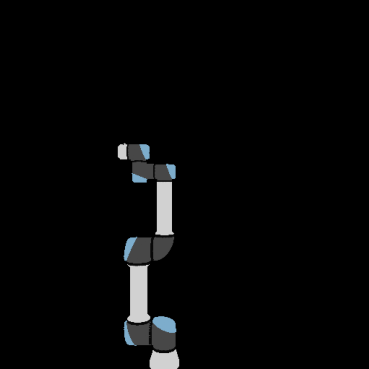}}
    \hspace{0.1mm}
    \subfloat[iter = 500]{\includegraphics[width=0.13\linewidth]{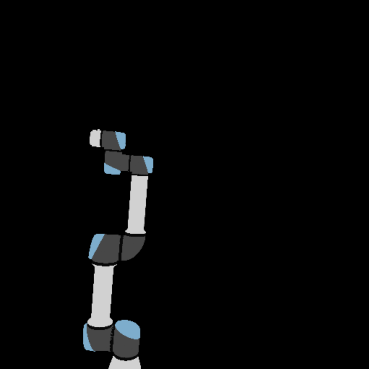}}

    \caption{Comparison of optimization processes using EasyHeC\cite{chen2023easyhec} (first row), IPE\cite{lu2023image} (second row), and DRHeC (third row) on the synthetic UR5e dataset. All methods start from the same initial pose and target ground-truth pose. EasyHeC and IPE struggle to converge when the initial pose deviates significantly from the ground truth due to large or incorrect gradients, often leading to optimization failure. In those methods, the manipulator flies out of the frame during the optimization process. In contrast, the proposed DRHeC method effectively mitigates these issues and successfully converges.}
    \label{fig:optimization}
\end{figure*}

The marker-based method tracks the relative motions of the camera and the robot's end-effector, forming a loop closure equation to estimate the hand-eye transformation matrix. The commonly used formulation is:
\begin{equation}
    AX = XB,
    \label{eq:handeye calibration}
\end{equation}
where $A$ and $B$ represent the robot's end-effector motion and the camera motion, respectively, and $X$ is the desired hand-eye transformation matrix.

Depending on the robot setup, this equation can be adapted to account for additional coordinate transformations or external reference frames, such as $AX=YB$, $AXB=YCZ$. Methods for solving the equation are typically classified into separate and simultaneous approaches \cite{park1994robot, horaud1995hand, andreff1999line}. Tsai et al. first compute the rotation using SVD, then estimate the translation using least squares\cite{tsai1989new}, while Daniilidis uses dual quaternions to solve both rotation and translation simultaneously, offering greater robustness to noise \cite{daniilidis1999hand}. However, the precision of these methods depends on marker accuracy, measurement poses, computational precision, and this process needs to be carried out before a manipulation task with a prepared marker.

Learning-based methods leverage deep learning to estimate the hand-eye transformation from the image. Lee et al. apply a Perspective-n-Point (PnP) algorithm to extract key points and compute the transformation \cite{lee2020camera, lu2022pose}, while Valassakis et al. propose a neural network-based direct regression approach \cite{valassakis2022learning}. However, these approaches rely on large-scale, high-quality datasets, which are costly to obtain and often lack transparency.

Recently, differentiable rendering-based hand-eye calibration has emerged as a promising alternative.
These methods optimize the hand-eye transformation by comparing rendered and real images, computing gradients through a differentiable pipeline.
While effective, existing differentiable rendering approaches face convergence issues and optimization errors, especially under limited binary-mask supervision.
In the next section, we discuss the challenges of current differentiable rendering-based methods and introduce our improvements.

\subsection{Differentiable Renderer Hand-eye Calibration}
Forward rendering involves generating a 2D image from a 3D model by considering various scene parameters such as shapes, materials, object poses, and lighting.
Differentiable rendering extends the rendering process by enabling the computation of the gradient between the 2D image and the 3D scene, facilitating optimization tasks such as pose estimation.
This technique has recently gained significant attention due to its ability to optimize 3D parameters in an end-to-end differentiable manner, making it a valuable tool in numerous applications \cite{mildenhall2021nerf, kerbl20233d, rosinol2023nerf, matsuki2024gaussian}.
One notable application is pose estimation, where differentiable rendering compares a rendered image with a real-world image to calculate the pose error.
By iteratively optimizing the pose parameters, the rendered image progressively aligns with the real image, achieving precise pose estimation \cite{palazzi2018end, park2020latentfusion}.

Recent studies explore the use of differentiable rendering specifically for pose estimation in robotics.
For example, Lu et al. proposed a method for precise pose estimation by employing differentiable rendering with distance maps and appearance differences, as well as reconstructing robot shapes from images using differentiable rendering in conjunction with a keypoint detector \cite{lu2023image}.
Additionally, EasyHeC proposes an approach for precise and automatic hand-eye calibration by combining differentiable rendering with space exploration techniques to optimize the calibration process \cite{chen2023easyhec,hong2024easyhec++}.

These methods depend on binary masks for differentiable rendering. However, binary masks often lose internal profile details, resulting in reduced precision and stability, and may also lead to pose ambiguity in some cases.
The complexity of robotic models makes automatic differentiation difficult, often leading to inaccurate gradient estimations and suboptimal convergence, as is shown in Fig~\ref{fig:optimization}.
To address these issues, we propose a novel approach that leverages RGB derivatives and mask geometric features such as centroid and area to enhance both robustness and accuracy.

\subsection{Image-to-Image Translation}
Image-to-Image Translation (I2IT) maps images between source and target domains, evolving from early classification- and regression-based methods \cite{long2015fully, xie2015holistically, zhang2016colorful}. The advent of Generative Adversarial Networks (GANs) significantly advanced this field, enabling realistic image synthesis \cite{goodfellow2014generative}. Conditional GANs, such as Pix2Pix and Pix2PixHD, further improved control over generation by leveraging paired training data \cite{pathak2016context, li2016precomputed, mirza2014conditional, isola2017image, wang2018high}. However, acquiring such paired datasets remains challenging.

CycleGAN addresses this limitation by introducing cycle consistency loss, enabling unpaired image translation by enforcing bidirectional consistency \cite{zhu2017unpaired}. Recent works extend CycleGAN across various aspects \cite{park2020contrastive}, such as GeoMaskGAN, which incorporates segmentation masks to improve geometric consistency in I2IT \cite{lu2022geometry}. While recent studies have mainly used diffusion models for I2IT \cite{ho2020denoising, song2020denoising, li2023bbdm}, CycleGAN has been shown to be more effective than diffusion models when dealing with unpaired datasets. Our study focuses on translating robotic images from real to simulation domains, where discrepancies in geometric sizes pose challenges. To mitigate this, we enhance CycleGAN with a mask loss to reduce geometric inconsistencies and improve translation accuracy.

\begin{figure*}[t]
  \centering
  \includegraphics[width=\textwidth]{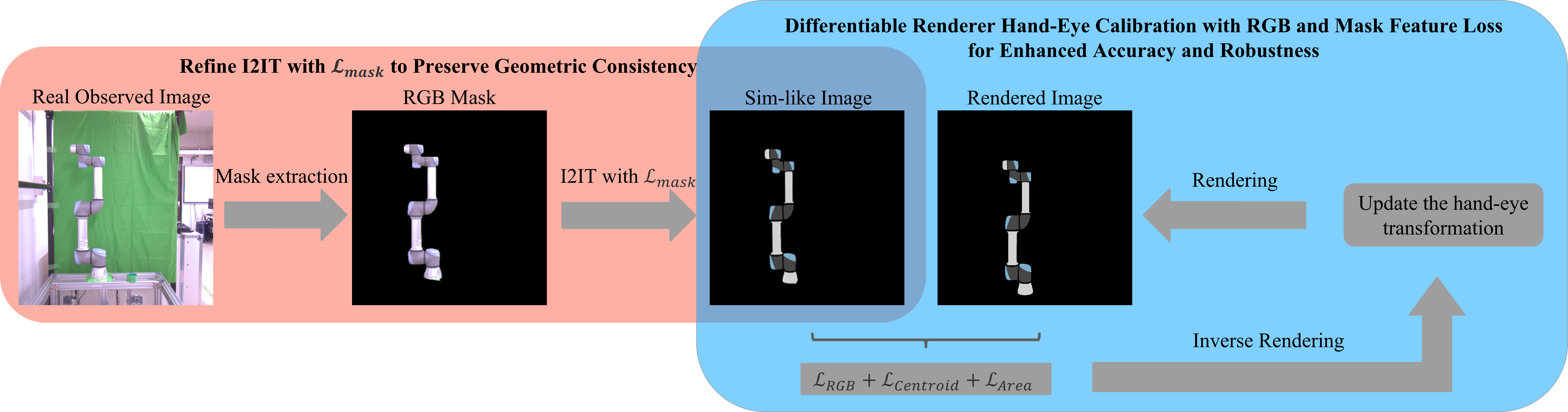}
  \caption{The architecture of the differentiable renderer hand-eye calibration. The observed real image is first transformed into a simulation-like (sim-like) image using I2IT, which is then compared with the rendered image for differentiable renderer hand-eye calibration.}
  \label{fig:differentiable rendering}
\end{figure*}

\section{Preliminaries}
\label{sec:Preliminaries}
Before presenting our method, we first review the foundational concepts of differentiable rendering.

\subsection{Differentiable Rendering}
The forward rendering process is defined as:
\begin{equation}
    I=f(\rho),
    \label{eq:forward rendering}
\end{equation}
where $I$ represents the rendered image, $f$ defines the mapping from the 3D model to the 2D image, and $\rho$ denotes the scene parameters.
By differentiating Eq.~(\ref{eq:forward rendering}) with respect to the scene parameters $\rho$, we have:
\begin{equation}
    \frac{\partial I}{\partial \rho} = \frac{\partial f(\rho)}{\partial \rho}.
    \label{eq:derivative}
\end{equation}
In inverse rendering, the goal is to recover the optimal scene parameters
$\rho$ by minimizing the discrepancy between the rendered image $f(\rho)$ and the observed image $I_{\text{obs}}$. This can be formulated as the following optimization problem:
\begin{equation}
    \hat{\rho} = \arg\min_{\rho} \mathcal{L}^{(2)}(f(\rho), I_{\text{obs}}).
    \label{eq:inverse rendering}
\end{equation}
where $\mathcal{L}^{(2)}$ denotes the pixel-wise squared error between the rendered image and the observed image.

\subsection{Differentiable Renderer Hand-Eye Calibration}
For single-image hand-eye calibration, the EasyHeC\cite{chen2023easyhec} method can be simplified to differentiable rendering using a single binary image. Assuming that the robot model is known, and the robot base pose relative to the camera frame can be determined based on an initial estimate of the hand-eye transformation, which is roughly obtained through manual measurement or approximate calibration. Given this information, the binary image of the robot can be rendered as:
\begin{equation}
    I_b = \mathcal{B}(f(q, {^{c}T_{b}})),
\end{equation}
where $I_b$ is the binary rendered image, $\mathcal{B}$ is the binary mask extraction operator, and $f$ is the forward rendering function that maps the robot model to the image. $q$ and ${^{c}T_{b}}$ represent the scene parameters, with $q$ denoting the joint angles that define the robot configuration and ${^{c}T_{b}}$ representing the hand-eye transformation.
In our formulation, the robot joint configuration $q$ is assumed to be known from the robot state and is not estimated during hand-eye calibration.
Accordingly, the unknown variable to be optimized is the hand-eye transformation ${^{c}T_{b}}$.
The proposed differentiable renderer calibration framework does not require exact appearance consistency between the real robot and the simulation model, since the I2IT module is introduced to reduce the real-to-sim appearance gap.
However, the robot geometry is still assumed to be accurate enough for the rendered observations to remain meaningfully aligned with the sim-like images.

The desired estimate of the hand-eye transformation can then be obtained by solving the following optimization problem until convergence:
\begin{equation}
    \hat{^{c}T_{b}} = \arg\min_{^{c}T_{b}} \mathcal{L}_2 \left(I_b, \mathcal{B}(I_{\text{obs}})\right).
\end{equation}

To minimize the loss function, we employ the Adam\cite{kingma2014adam} optimizer. By iteratively updating $^{c}T_{b}$, the rendered binary mask $I_b$ is gradually aligned with the observed binary mask $\mathcal{B}(I_{\text{obs}})$, leading to an accurate estimate of the hand-eye transformation.

While EasyHeC is effective in many cases, it still faces several challenges. For instance, the lack of global gradient direction can lead to instability in the optimization process, causing vanishing or exploding gradients, which can even result in the rendered mask moving out of the frame. Additionally, since binary masks lack color information, the calibration precision is limited and may result in pose ambiguity in some cases.

\section{Method}
\label{sec:Method}

To overcome these issues, we reformulate the calibration objective by combining local RGB-based appearance information with global mask-geometric constraints.
First, RGB derivatives provide dense local appearance cues that improve precision and reduce the risk of local minima compared with binary-mask-based supervision.
Second, centroid and area losses provide global geometric guidance for the rendered robot mask, which stabilizes the optimization when the initial pose error is large.
In this sense, the proposed method does not simply add image features, but provides more reliable gradient guidance for stable optimization.

The proposed framework consists of two main components: RGB-based differentiable-renderer hand-eye calibration and geometry-preserving real-to-simulation I2IT.
Compared with EasyHeC\cite{chen2023easyhec} and IPE\cite{lu2023image}, which mainly rely on silhouette cues derived from binary masks, the proposed framework additionally incorporates RGB appearance information and both mask centroid and area constraints to improve the stability of differentiable-rendering optimization.
For the I2IT module, our method builds upon CycleGAN and introduces a mask loss to reduce geometric inconsistencies during image translation.
Together, these components form a unified RGB-based differentiable-rendering framework for hand-eye calibration.

\subsection{I2IT}
\label{sec:I2IT}
To leverage RGB derivatives as differentiable parameters, it is essential to compute the loss between the RGB mask and the rendered image. However, these two images often exhibit significant differences due to factors such as variations in model size, color, material properties, and lighting conditions. These differences arise from the variations between the real robot and the simulated model, making it challenging to directly compare the two images.

A feasible solution to address this challenge is the I2IT method, which transforms real-world robot images into simulation-like images, which are images that resemble those generated by rendering in simulation, as shown in the red section on the left of Fig.~\ref{fig:differentiable rendering}. First, the RGB mask is extracted from the real observed image, and then an I2IT transformation is applied to convert the RGB mask into a simulation-like image. However, it is well-known that no existing I2IT model can be generally applied across different types of robot manipulators for real-to-sim image translation. Consequently, a robot-specific I2IT model must be trained for each specific application. By minimizing the disparities between the rendered and real images, particularly when both share the same hand-eye transformation and robot configuration, this method enhances the robustness of the optimization process and improves the accuracy of the RGB loss calculation. In this work, we refine the CycleGAN method, where the generator's loss function in the standard CycleGAN framework is defined as:
\begin{equation}
    \mathcal{L} = \lambda_{\text{id}}\mathcal{L}_{\text{id}}^{(1)} + \lambda_{\text{GAN}}\mathcal{L}_{\text{GAN}} + \lambda_{\text{cycle}}\mathcal{L}_{\text{cycle}}^{(1)},
\end{equation}
where $\mathcal{L}_{\text{id}}$ represents the identity loss, ensuring that when images from the target domain are fed into the generator, their domain-specific features are preserved without unnecessary changes. $\mathcal{L}_{\text{GAN}}$ denotes the adversarial loss, which ensures that the generated images are indistinguishable from real images in the target domain, while $\mathcal{L}_{\text{cycle}}$ is the cycle consistency loss, which guarantees that an image can be reconstructed to its original form after being translated to the target domain and then back to the source domain. The coefficients $\lambda_{\text{id}}$, $\lambda_{\text{GAN}}$, and $\lambda_{\text{cycle}}$ serve as the weighting factors for each loss term.

Despite the I2IT process, minor pose misalignments may still arise due to configuration discrepancies between the real and simulation robot models.
These misalignments can introduce errors in subsequent hand-eye calibration.
To mitigate such effects, we introduce two practical measures for the I2IT training stage:
(1) an offline preliminary hand-eye calibration with binary images to improve geometric alignment between the real and simulated models when preparing the I2IT training pairs, and
(2) a mask loss term in the CycleGAN training process to enhance geometric consistency.
The preliminary calibration is used only as a coarse data-alignment step for I2IT training, and it is not used to initialize or constrain the final DRHeC optimization.
After the I2IT network is trained, its parameters are fixed, and no final DRHeC calibration result is fed back to retrain or update the I2IT module.
In our current implementation, the pretrained I2IT module provides real-to-simulation appearance alignment while maintaining geometric consistency before RGB-based differentiable-renderer hand-eye calibration.
The revised loss function is expressed as:
\begin{equation}
    \mathcal{L} = \lambda_{\text{id}}\mathcal{L}_{\text{id}}^{(1)} + \lambda_{\text{GAN}}\mathcal{L}_{\text{GAN}} + \lambda_{\text{cycle}}\mathcal{L}_{\text{cycle}}^{(1)} + \lambda_{\text{mask}}\mathcal{L}_{\text{mask}}^{(1)}.
\end{equation}
where the mask loss $\mathcal{L}_{\text{mask}}^{(1)} = \mathbb{E}[\|\mathcal{B}(G(x)) - \mathcal{B}(x)\|_1]$ penalizes discrepancies between the binary masks of the generated images and those of the inputs. $\lambda_{\text{mask}}$ is a weighting factor that controls the contribution of the mask loss.

These improvements facilitate better geometric and color consistency, ensuring more accurate and reliable image translations between the real and simulated domains.

\begin{algorithm}[!t]
\caption{Hand-Eye Calibration via Differentiable Rendering with RGB and Mask Geometry Features}
\textbf{Input:} Observed image \(I_{ref}\), initial hand-eye pose \(q_{init}\)\\
\textbf{Output:} Optimized hand-eye pose \(q_{opt}\)

\begin{algorithmic}[1]
\STATE \(q_{opt} \gets q_{init}\)
\STATE optimizer \(\gets\) Adam(\([q_{opt}]\), \(lr_{base}\))
\FOR{\(i = 1\) to \(iter_{max}\)}
    \STATE \(I_{opt} \gets f(q_{opt})\)
    \STATE \(L_{rgb} \gets \text{MSE}(I_{ref}, I_{opt})\)
    \STATE \(L_{centroid} \gets \text{centroid\_loss}(\mathcal{B}(I_{ref}), \mathcal{B}(I_{opt}))\)
    \STATE \(L_{area} \gets \text{area\_loss}(\mathcal{B}(I_{ref}), \mathcal{B}(I_{opt}))\)
    \IF{\(L_{centroid} > \text{threshold}\)}
        \STATE \(L_{total} \gets L_{rgb} + \lambda_c L_{centroid} + \lambda_a L_{area}\)
    \ELSE
        \STATE \(L_{total} \gets L_{rgb}\)
    \ENDIF
    \STATE optimizer.zero\_grad()
    \STATE \(L_{total}\).backward()
    \IF{\(L_{centroid} > \text{threshold}\)}
        \STATE \textbf{with} \texttt{torch.no\_grad()} \textbf{do}
        \STATE \quad \(\nabla q_{opt}[3:] = 0\) \COMMENT{Freeze rotation gradient}
    \ENDIF
    \STATE optimizer.step()
\ENDFOR
\STATE \textbf{return} \(q_{opt}\)
\end{algorithmic}
\label{algorithm:calibration}
\end{algorithm}

\subsection{RGB-Based Differentiable Rendering}
As shown in Fig.~\ref{fig:differentiable rendering}, after transforming the real observed image into a simulation-like image, we can compare it with the rendered simulation image to infer the hand-eye transformation. To improve the precision of the hand-eye calibration, we initially attempt to use RGB derivatives instead of binary derivatives during calibration, as described by the following equation:
\begin{equation}
    \hat{^{c}T_{b}} = \arg\min_{^{c}T_{b}} \mathcal{L}^{(2)} \left(I_{rgb}, I_{\text{sim-like}}\right),
\end{equation}
where $ I_{rgb} $ is the rendered RGB image. $ I_{\text{sim-like}} $ is the simulation-like image obtained via the I2IT transformation.

However, based on experimental observations, we found that this approach does not yield satisfactory results. The use of RGB derivatives introduces more detailed color information, which leads to uneven gradient propagation compared to binary masks. This unevenness can cause the optimization to diverge and increase the likelihood of the solution going out of frame during differentiable rendering. To address this issue, we propose a novel method that introduces a loss function aimed at capturing the global gradient direction with respect to the robot mask. By combining the original RGB derivatives loss with the centroid distance loss and mask area loss, we aim to improve both calibration precision and optimization stability. The revised equation is shown as:
\begin{equation}
    \label{eq:loss_function}
    \begin{aligned}
        \hat{^{c}T_{b}} = \arg\min_{^{c}T_{b}} \Bigg( & \lambda_{\text{rgb}} \mathcal{L}^{(2)} \left(I_{rgb}, I_{\text{sim-like}}\right) \\
        & + \lambda_{\text{centroid}} \mathcal{L}_{\text{centroid}}^{(2)} \left(I_{rgb}, I_{\text{sim-like}}\right) \\
        & + \lambda_{\text{area}} \mathcal{L}_{\text{area}}^{(2)} \left(I_{rgb}, I_{\text{sim-like}}\right) \Bigg).
    \end{aligned}
\end{equation}
where $\mathcal{L}^{(2)} \left(I_{rgb}, I_{\text{sim-like}}\right)$ calculates the RGB loss between the rendered and the simulation-like images. $\mathcal{L}_{\text{centroid}}^{(2)}$ represents the centroid distance loss between the binary masks of the two images, and $\mathcal{L}_{\text{area}}^{(2)}$ computes their area loss. $\lambda_{\text{rgb}}$, $\lambda_{\text{centroid}}$, and $\lambda_{\text{area}}$ define the weighting factors for each loss.

The centroid loss quantifies the Euclidean distance between soft-weighted centroids of the rendered and simulation-like images:
\begin{equation}
\mathcal{L}_{\text{centroid}}^{(2)} (I_{rgb}, I_{\text{sim-like}}) = \left\| \mathbf{c}(I_{rgb}) - \mathbf{c}(I_{\text{sim-like}}) \right\|_2,
\end{equation}
where $\mathbf{c}(\cdot)$ denotes the soft centroid derived from the grayscale intensity, computed as the weighted average of pixel coordinates:
\begin{equation}
\mathbf{c}(I) = \sum\limits_{p \in P} \frac{\exp\left(\mathcal{G}(I)[p] / \tau\right)}{\sum\limits_{r \in P} \exp\left(\mathcal{G}(I)[r] / \tau\right)} \cdot \mathbf{x}_p,
\end{equation}
where $P$ denotes the set of all pixel indices in the image, and $p, r \in P$ serve as index variables. The vector $\mathbf{x}_p$ is the 2D image coordinate of pixel $p$. The denominator computes a softmax normalization over all pixels, ensuring that the weights sum to 1. $\tau$ is a temperature parameter, typically set to 1.0.
\begin{equation}
\mathcal{G}(I)[p] = \frac{1}{K} \sum_{k=1}^K I[p, k].
\end{equation}
where $K=3$ is the number of RGB channels, and $k = 1, 2, 3$ indexes these color channels.

Compared to binary masks that assign equal weights to all object pixels and have hard edges causing non-smooth, non-differentiable boundaries, soft weights provide a smooth, fully differentiable way to measure differences. This smoothness helps the centroid change continuously with small image changes, improving gradient flow during optimization and leading to more stable and accurate results.

The area loss is defined based on a smooth and differentiable approximation of the object area computed from the grayscale image.
\begin{equation}
\mathbf{a}(I) = \sum_{p} \sigma\big(\mathcal{G}(I)[p]\big),
\end{equation}
where \(\sigma(\cdot)\) is the sigmoid function and \(\mathcal{G}(I)[p]\) denotes the grayscale intensity at pixel \(p\).
Then area loss can be defined as:
\begin{equation}
\mathcal{L}_{\text{area}}^{(2)} (I_{rgb}, I_{\text{sim-like}}) = \left\| \mathbf{a}(I_{rgb}) - \mathbf{a}(I_{\text{sim-like}}) \right\|_2.
\end{equation}

In practice, we observed that during early optimization stages, large pose errors can make the calibration objective under-constrained and poorly conditioned, which can lead to divergence or poor convergence of the hand-eye pose estimation.
To address this, we adopt a two-step optimization strategy:
Initially, we freeze the gradients of the three Euler rotation variables when the centroid loss exceeds a threshold, indicating that the current estimate is still in the coarse-alignment stage, effectively restricting updates to the translation parameters.
This allows the pose to move closer to the true position under the guidance of our improved combined loss function, stabilizing the optimization.
Once the centroid loss falls below a predefined switching threshold, these three rotation gradients are unfrozen for fine-tuning using the full RGB loss, marking the transition to the fine-refinement stage and enabling precise rotation refinement.
This staged approach improves stability and accuracy by preventing erratic updates in rotation during early optimization.
The switching threshold controls the trade-off between optimization stability and refinement efficiency. If the threshold is too small, rotation updates are enabled only after the centroid misalignment becomes very small, which improves early-stage stability but may delay rotation refinement and slow convergence. If the threshold is too large, rotation updates are enabled earlier, which may accelerate refinement but can also expose the optimization to noisy or misleading rotation gradients when the pose is still poorly aligned.
Since the calibration module does not feed its outputs back into the I2IT module, the overall pipeline does not introduce recursive feedback between I2IT training and hand-eye calibration.
This approach effectively stabilizes early optimization and improves final accuracy. Pseudocode for this procedure is provided in Algorithm~\ref{algorithm:calibration}.

In contrast to the IPE method \cite{lu2023image}, which relies on the rendered image mask multiplied by a distance map, this results in a non-smooth loss function. To address this issue, the gradient at the edge requires special processing \cite{li2018differentiable}, but the current implementation still faces gradient issues.
Moreover, for non-convex robot configurations, the gradient directions can become inconsistent, further hindering convergence.
To overcome these issues, our method incorporates a differentiable mask centroid loss that facilitates smoother gradient propagation and is more robust to shape complexity.
However, while the centroid loss effectively aligns the mask centers, it may not always precisely correspond to the object's true center, potentially resulting in misalignment along the z-axis.
To address this, we incorporate an area loss to stabilize z-translation by enforcing mask area consistency.

\subsection{Implementation Details}
The key hyperparameters in our optimization process were determined through preliminary small-scale experiments to ensure stable convergence and reliable performance.
For the I2IT training stage, we follow the original CycleGAN settings by keeping the weights $\lambda_{\text{id}}$, $\lambda_{\text{GAN}}$, and $\lambda_{\text{cycle}}$ unchanged, and set $\lambda_{\text{mask}} = 1.0$.
The network is trained for 400 epochs, with a constant learning rate for the first 200 epochs, followed by a linear decay over the remaining epochs.
For the differentiable rendering-based optimization, we use the Adam optimizer with a fixed learning rate of 0.002 and a maximum of 2000 iterations.
The loss weights are set as follows: $\lambda_{\text{rgb}} = 1.0$, $\lambda_{\text{centroid}} = 0.01$, and $\lambda_{\text{area}} = 0.0001$.
These values were chosen to reflect the intended roles of each loss component: the RGB loss functions as the dominant optimization signal, the centroid distance loss introduces geometric regularization, and the mask area loss enforces additional shape consistency.
We set the threshold for the two-step optimization to 0.25.
This value serves as a practical criterion for controlling the transition from coarse translation-dominant alignment to full pose refinement. As shown in the ablation study, the method performs consistently within a nearby range of threshold values, indicating that the optimization is not strongly dependent on this exact value.
Note that the threshold and loss weights may require tuning for substantially different environments or datasets.
Imbalanced settings could lead to optimization instability or suboptimal convergence, highlighting the importance of appropriate loss balancing in the rendering-based calibration process.

\section{Simulations}
\begin{figure}[t]
  \centering
  \subfloat[]{\includegraphics[width=\linewidth]{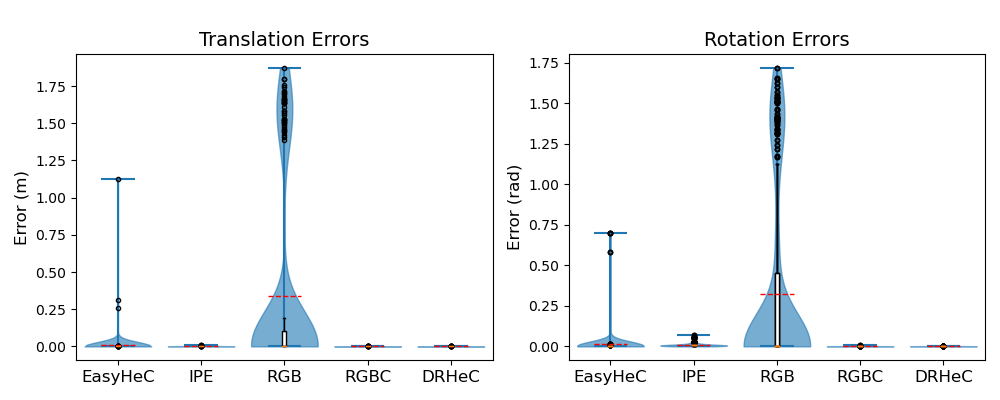} \label{fig:exp_LP}} \\
  \subfloat[]{\includegraphics[width=\linewidth]{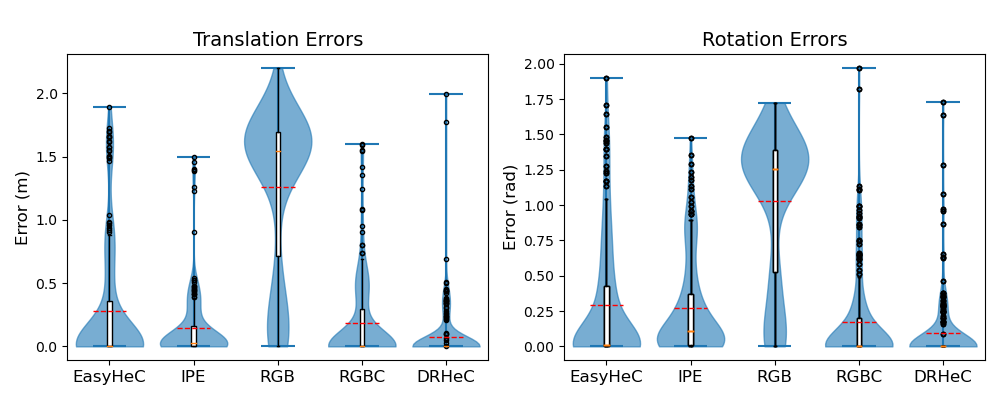} \label{fig:exp_MP}} \\
  \subfloat[]{\includegraphics[width=\linewidth]{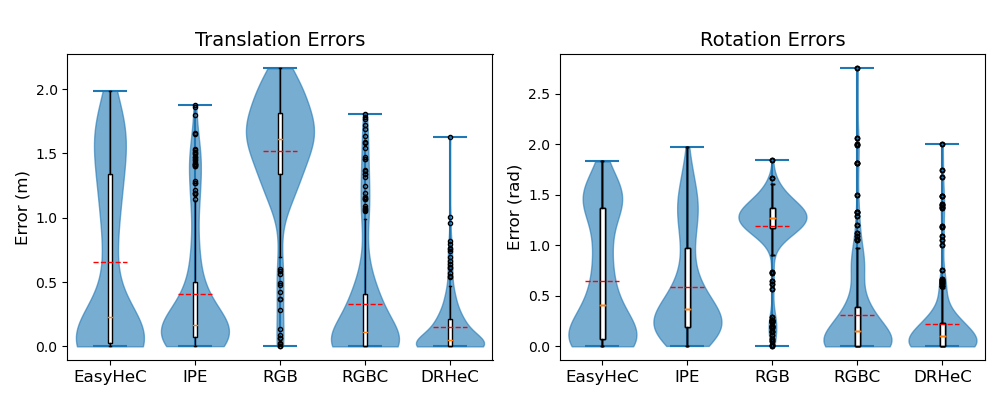} \label{fig:exp_HP}}
  \caption{Distributions of calibration errors under varying levels of initial perturbations for different differentiable rendering methods: EasyHeC\cite{chen2023easyhec}, IPE\cite{lu2023image}, RGB, RGBC (RGB and Mask Centroid), and DRHeC. (a) Low perturbations, (b) Moderate perturbations, (c) High perturbations.}
  \label{fig:calibration_errors}
\end{figure}

To evaluate the effectiveness of the proposed method, we conduct simulation experiments on both synthetic and real-world datasets. The synthetic dataset allows us to compare our approach with differentiable rendering-based methods under various initial pose errors, while the real-world dataset enables direct evaluation against learning-based hand-eye calibration approaches.

\subsection{Synthetic Datasets}
\label{subsec:Synthetic Dataset}

\begin{table*}[ht]
\caption{Experimental Results on the Synthetic UR5e Dataset.}
\label{table:synthetic_sim_result}
\centering
\begin{tabular}{l|l|cccc|cccc|c}
\hline
\multirow{2}{*}{Perturbation Level} & \multirow{2}{*}{Method} & \multicolumn{4}{c|}{Translation Error (meters)} & \multicolumn{4}{c|}{Rotation Error (radians)} & \multirow{2}{*}{Success Rate (\%)} \\ \cline{3-10}
& & Q1 & Q3 & Max & Mean & Q1 & Q3 & Max & Mean & \\
\hline
\multirow{5}{*}{Low (LP)} & EasyHeC \cite{chen2023easyhec} & 0.0002 & 0.0005 & 1.1282 & 0.0089 & 0.0010 & 0.0029 & 0.7012 & 0.0122 & 91.5 \\
& IPE \cite{lu2023image} & 0.0008 & 0.0021 & 0.0084 & 0.0016 & 0.0034 & 0.0100 & 0.0693 & 0.0095 & \textbf{100.0} \\
& RGB & 0.0002 & 0.1037 & 1.8703 & 0.3419 & 0.0008 & 0.4508 & 1.7163 & 0.3254 & 79.0 \\
& RGB+Mask Centroid & \textbf{0.0001} & \textbf{0.0003} & \textbf{0.0012} & \textbf{0.0003} & 0.0007 & 0.0016 & 0.0080 & 0.0013 & \textbf{100.0} \\
& DRHeC & 0.0002 & \textbf{0.0003} & 0.0013 & \textbf{0.0003} & \textbf{0.0006} & \textbf{0.0015} & \textbf{0.0053} & \textbf{0.0012} & \textbf{100.0} \\
\hline
\multirow{5}{*}{Medium (MP)} & EasyHeC \cite{chen2023easyhec} & 0.0005 & 0.3570 & 1.8940 & 0.2815 & 0.0016 & 0.4274 & 1.8958 & 0.2899 & 91.5 \\
& IPE \cite{lu2023image} & 0.0034 & 0.1573 & \textbf{1.4977} & 0.1469 & 0.0121 & 0.3683 & \textbf{1.4743} & 0.2719 & 96.5 \\
& RGB & 0.7112 & 1.6945 & 2.1994 & 1.2572 & 0.5230 & 1.3933 & 1.7245 & 1.0278 & 28.5 \\
& RGB+Mask Centroid & \textbf{0.0002} & 0.2933 & 1.5972 & 0.1862 & 0.0007 & 0.2041 & 1.9684 & 0.1703 & \textbf{99.0} \\
& DRHeC & \textbf{0.0002} & \textbf{0.0009} & 1.9937 & \textbf{0.0766} & \textbf{0.0007} & \textbf{0.0031} & 1.7313 & \textbf{0.0930} & \textbf{99.0} \\
\hline
\multirow{5}{*}{High (HP)} & EasyHeC \cite{chen2023easyhec} & 0.0293 & 1.3432 & 1.9862 & 0.6572 & 0.0709 & 1.3728 & \textbf{1.8319} & 0.6508 & 71.0 \\
& IPE \cite{lu2023image} & 0.0729 & 0.4965 & 1.8722 & 0.4055 & 0.1910 & 0.9723 & 1.9769 & 0.5866 & 83.5 \\
& RGB & 1.3399 & 1.8105 & 2.1614 & 1.5204 & 1.1760 & 1.3681 & 1.8452 & 1.1892 & 11.0 \\
& RGB+Mask Centroid & \textbf{0.0005} & 0.4068 & 1.8095 & 0.3323 & 0.0012 & 0.3923 & 2.7546 & 0.3124 & 97.5 \\
& DRHeC & \textbf{0.0005} & \textbf{0.2103} & \textbf{1.6269} & \textbf{0.1495} & \textbf{0.0011} & \textbf{0.2351} & 2.0007 & \textbf{0.2187} & \textbf{99.5} \\
\hline
\end{tabular}
\end{table*}

In this subsection, we conduct studies on the synthetic UR5e dataset to evaluate the individual contributions of key components in our proposed framework.
We generate a synthetic dataset using a UR5e robot to systematically evaluate our method under controlled conditions.
For each synthetic sample, the robot joint configuration is generated by sampling around the UR5e robot pose $[0, -90^\circ, 90^\circ, -90^\circ, -90^\circ, 0]$, with an independent perturbation of $\pm 10^\circ$ applied to each joint.

To ensure a comprehensive assessment, we generate 200 test samples for each perturbation level, randomly selected within predefined error ranges.
Three experimental settings are designed to simulate different levels of initial misalignment: (LP) low translation errors within $[-0.1,0.1]$ meters and rotation errors within $[- \pi/18, \pi/18]$ radians, (MP) moderate translation errors from $[-0.2,-0.1]$ to $[0.1,0.2]$ meters, and (HP) high translation errors from $[-0.3,-0.2]$ to $[0.2,0.3]$ meters, while the rotation error range remains fixed at $[- \pi/18, \pi/18]$ radians.
These settings allow us to evaluate the precision and robustness of each method under varying levels of misalignment.
We compare five rendering-based calibration methods: EasyHeC\cite{chen2023easyhec}, IPE\cite{lu2023image}, the RGB loss method, the RGB loss combined with centroid distance loss method (RGBC), and the DRHeC method.
Table~\ref{table:synthetic_sim_result} provides the quantitative results, while Figs.~\ref{fig:calibration_errors}\,(a), \ref{fig:calibration_errors}\,(b), and \ref{fig:calibration_errors}\,(c) visualize the corresponding outcomes.

Overall, the results demonstrate that adding the mask centroid loss to the RGB loss significantly reduces translation and rotation errors, confirming the importance of geometric constraints in guiding optimization. Further inclusion of the mask area loss in DRHeC yields additional improvements in error metrics and notably increases success rates, especially under challenging high-perturbation conditions. This confirms that leveraging both RGB-based gradients and mask-derived geometric information enhances the accuracy and robustness of hand-eye calibration.

More specifically, DRHeC consistently achieves the lowest Q3 (third quartile) and mean translation and rotation errors across all perturbation levels, demonstrating superior robustness to large initial misalignments compared to all baselines. In particular, under MP and HP settings, DRHeC consistently outperforms other methods, demonstrating superior robustness against large initial misalignments. Although the IPE method exhibits a slightly lower maximum rotation error in the MP setting, this can be attributed to the strong local constraints imposed by its distance map loss. However, this advantage is offset by its higher mean and Q3 values compared to DRHeC, indicating less consistent overall performance.

Moreover, the inclusion of mask area gradients further improves performance, as evidenced by the increase in success rates from 97.5\% (RGBC) to 99.5\% (DRHeC) in the HP setting. This result highlights the advantage of leveraging both RGB-based gradients and mask-derived geometric constraints, which enhances stability and accuracy.

\subsection{Real World Datasets}
To further validate our approach in real-world conditions, we use the real-world Baxter dataset from \cite{lu2022pose}. This dataset allows direct comparison with EasyHeC and other learning-based hand-eye calibration methods. It consists of 100 images captured from 20 distinct robot poses, with each image annotated with GT (Ground Truth) keypoints for quantitative evaluation.

Following the EasyHeC setup, we manually initialize the camera pose relative to the Baxter robot.
The collected images are used to construct I2IT training pairs through an offline coarse alignment procedure.
This alignment step is used only to prepare the I2IT training pairs and is not part of the final DRHeC optimization.
The trained I2IT model then transforms real images into a simulation-like style, reducing the appearance gap between real and simulated robot images while preserving geometric consistency.
Finally, we apply the DRHeC method to calibrate the hand-eye pose and evaluate the results.
Together with the UR5e and DENSO VS060 experiments presented later, these results provide initial, although not exhaustive, evidence of the applicability of the proposed framework across different robot morphologies, visual appearances, and data sources.

\begin{table}[htbp]
\caption{Evaluation results of 2D PCK on the real-world dataset.}
\label{table:2D_PCK}
\centering
\footnotesize
\resizebox{\columnwidth}{!}{%
\begin{tabular}{cccccccc}
\hline
Method & 20px & 30px & 40px & 50px & 100px & 150px & 200px \\
\hline
Dream \cite{lee2020camera}           & 0.16 & 0.23 & 0.29 & 0.33 & 0.52 & 0.62 & 0.64 \\
OK \cite{lu2022pose}              & 0.34 & 0.54 & 0.66 & 0.69 & 0.88 & 0.93 & 0.95 \\
\hline
IPE (box) \cite{lu2023image}      & -    & -    & -    & 0.65 & 0.94 & 0.95 & 0.95 \\
IPE (cylinder) \cite{lu2023image} & -    & -    & -    & 0.80 & 0.91 & 0.93 & 0.95 \\
IPE (CAD) \cite{lu2023image}      & -    & -    & -    & 0.74 & 0.90 & 0.95 & 0.95 \\
EasyHeC \cite{chen2023easyhec}    & 0.35 & 0.55 & 0.75 & 0.90 & 0.95 & 0.95 & \textbf{1.00} \\
EasyHeC++ \cite{hong2024easyhec++}    & 0.50 & \textbf{0.75} & 0.75 & 0.85 & 0.90 & 0.95 & \textbf{1.00} \\
DRHeC           & \textbf{0.65} & \textbf{0.75} & \textbf{0.90} & \textbf{1.00} & \textbf{1.00} & \textbf{1.00} & \textbf{1.00} \\
\hline
\end{tabular}
}
\end{table}

The evaluation employs the 2D PCK (Percentage of Correct Keypoints) metric, which quantifies the proportion of estimated keypoints that fall within a specified pixel distance from the GT keypoints. Results are evaluated across multiple pixel thresholds (from 20px to 200px) to offer a comprehensive view of performance at various accuracy levels. As shown in Table~\ref{table:2D_PCK}, our proposed method consistently outperforms both learning-based and the differentiable renderer hand-eye calibration methods.

Notably, DRHeC outperforms other methods across all pixel thresholds, suggesting that our method not only improves calibration accuracy but also enhances robustness to small perturbations in real-world data. This demonstrates that our method offers both superior accuracy and greater stability for hand-eye calibration.
\section{Real-world Experiments}
To validate the proposed method, we conduct two types of real-world experiments. We primarily evaluate our approach on the UR5e robot, as described in Sec.~\ref{subsec:Experiment Setup} through Sec.~\ref{subsec:Insertion Experiments}. Additionally, we test our method on a DENSO VS060 robot, which has a nearly colorless appearance with the entire robot being white except for its black end-effector, to further demonstrate the effectiveness of our approach in scenarios with minimal color information (Sec.~\ref{subsec:Denso Robot Experiments}). Finally, we present a discussion in Sec.~\ref{subsec:Discussion}.

\subsection{Experiment Setup}
\label{subsec:Experiment Setup}

\begin{figure}[t]
  \centering
  \subfloat[]{\includegraphics[width=0.50\linewidth]{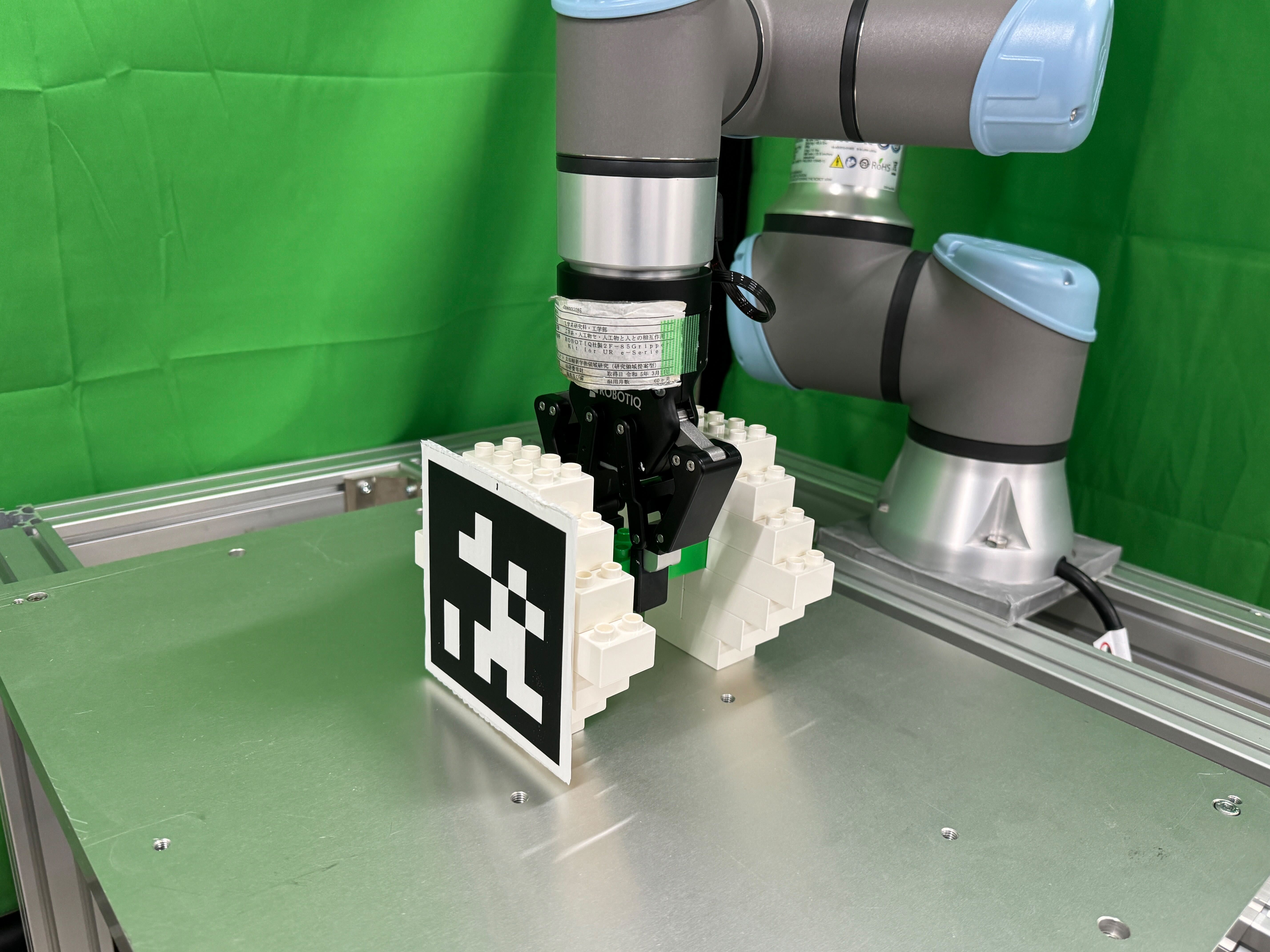} \label{fig:grasping}}
  \subfloat[]{\includegraphics[width=0.50\linewidth]{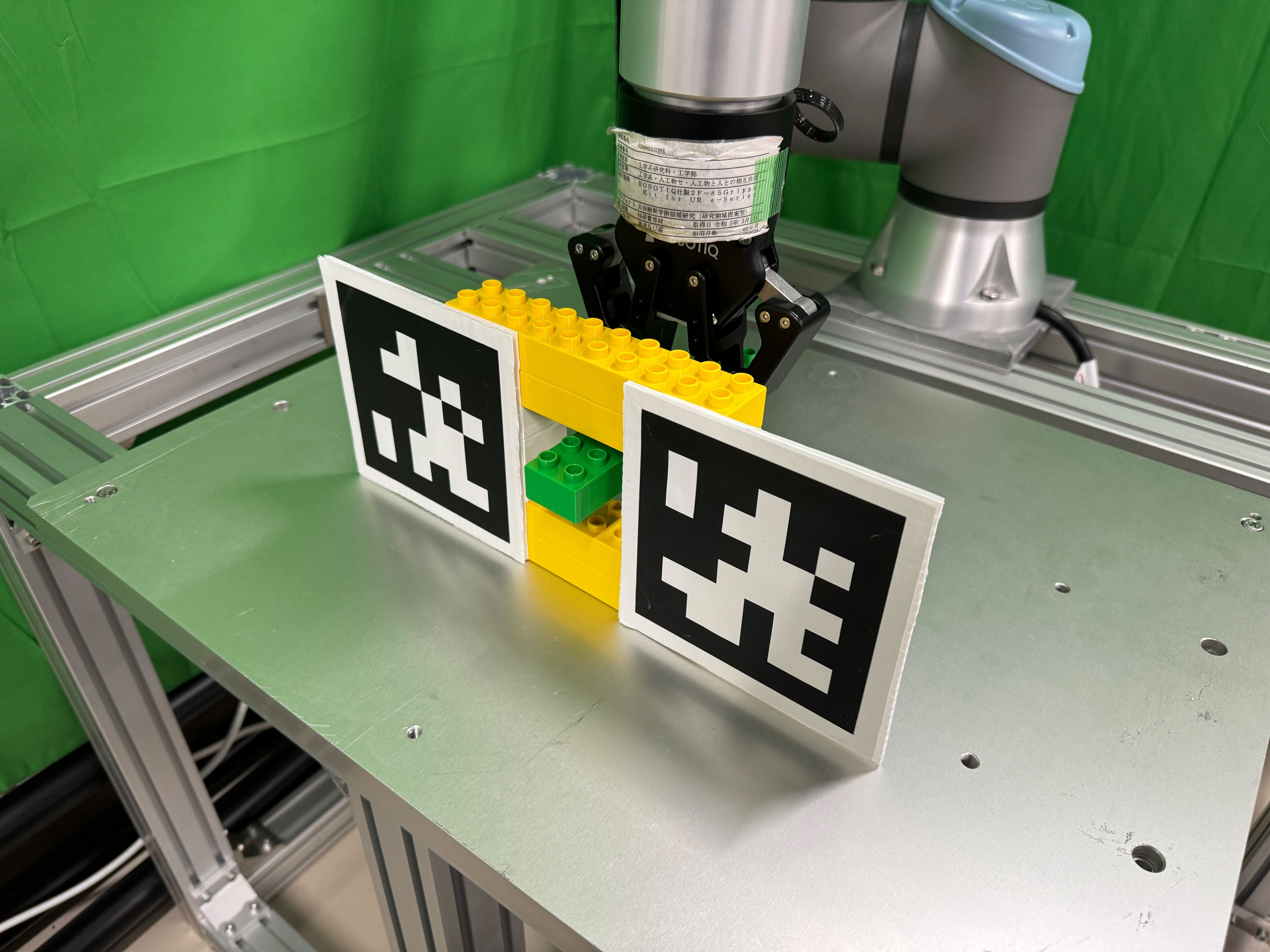} \label{fig:insertion}}
  \caption{Experimental setup for two robot tasks. (a) Grasping experiment to verify the robot's ability to grasp an object after hand-eye calibration. (b) Insertion experiment to verify the robot's ability to accurately insert into a hole after hand-eye calibration.}
  \label{fig:grasp_insertion}
\end{figure}

We conduct real-world experiments using a UR5e robot (Universal Robots A/S, Odense, Denmark). The experimental setup includes an industrial CMOS camera, the Basler acA2440-20gc (Basler AG, Schleswig-Holstein, Germany), equipped with a fixed-focus lens, the Kowa LM8JC10M (Kowa Optronics Co., Ltd., Aichi, Japan).
To ensure the camera can capture the entire UR5e robot within its field of view, the camera tripod is positioned approximately 2 meters away from the UR5e robot along the x-axis, with both remaining at roughly the same horizontal plane.
It is assumed that both the camera and the robot are well calibrated.
For differentiable rendering, we utilize nvdiffrast \cite{Laine2020diffrast} as the rendering engine, while image segmentation is performed using SAM \cite{kirillov2023segany}.
SAM is used only to generate the initial mask proposal, with manual prompt selection and mask correction when necessary, in order to ensure reliable masks and to prevent segmentation errors from influencing the evaluation of the proposed calibration framework.
This mask-preparation step is not part of the differentiable-renderer-based hand-eye optimization itself, and the segmentation module can be replaced by another task-specific automatic method without changing the proposed calibration framework.
All experiments are conducted on an NVIDIA RTX 4090 GPU.
All real-world experiments were performed under stable indoor lighting, with the workspace illuminated by fixed overhead LED lights, ensuring approximately constant illumination across all trials.

For this study, we selected two representative tasks that highlight key capabilities required for robotic manipulation: grasping and insertion, as shown in Fig.~\ref{fig:grasp_insertion}.
These tasks are sensitive to hand-eye calibration accuracy.
Grasping requires precise positioning to avoid missing the object, while insertion demands accurate alignment.
Thus, both serve as effective benchmarks for evaluating calibration performance.
To evaluate the proposed method, we compared it against several baseline approaches,
including two classical marker-based methods (Tsai's \cite{tsai1989new} and Daniilidis' \cite{daniilidis1999hand}),
which are widely used in industrial settings, as well as two recent marker-based methods, Single3D \cite{jin2024hand}
and Wang et al. \cite{wang2024dual}, for a more comprehensive academic comparison. We also include the
differentiable rendering method EasyHeC \cite{chen2023easyhec} as a baseline.
In both tasks, the object pose relative to the camera frame was estimated using 12 cm AprilTags and the PnP algorithm.
These AprilTags were used only as an auxiliary evaluation setup for object pose estimation in the downstream validation experiments, and this evaluation setup was applied uniformly to all compared methods. They were not used in the proposed differentiable-renderer-based hand-eye calibration procedure.
Based on this pose estimation, the gripper then calculated the object's position relative to the robot's base frame.
Notably, in the grasping experiments, object collision was not considered, making precise hand-eye calibration essential for achieving successful grasps.

\subsection{I2IT Preprocessing}
\label{subsec:I2IT Preprocessing}

\begin{figure}[tbp]
\centering
\includegraphics[width=\linewidth]{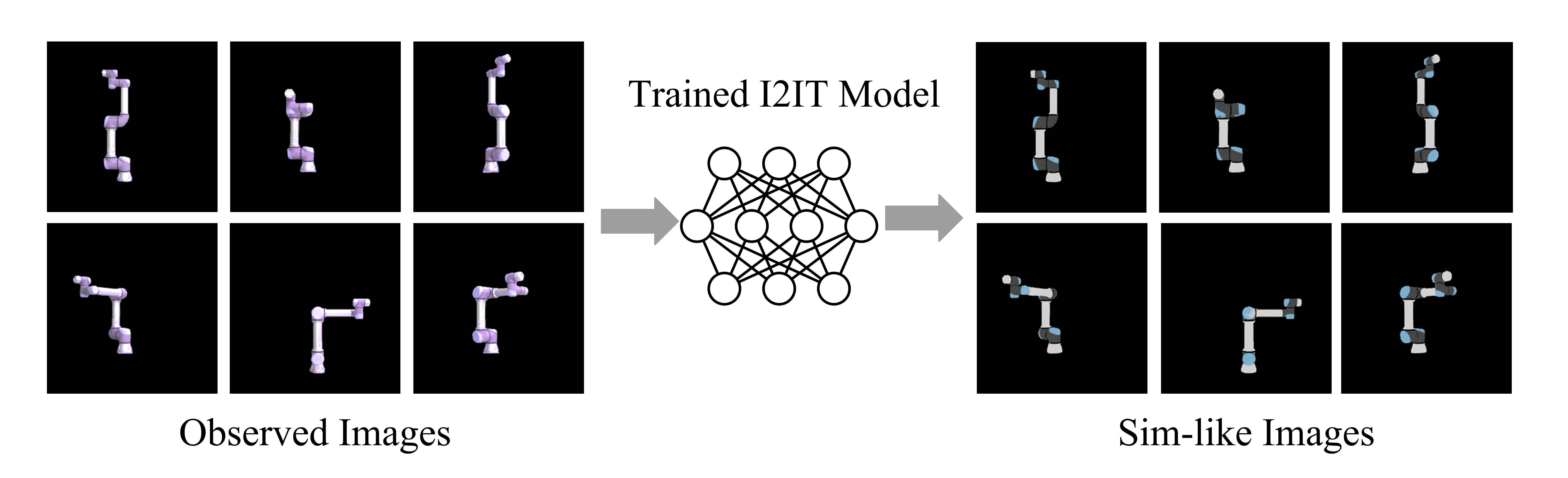}
\caption{Example of real-world image translated into a simulation-like image for differentiable rendering-based hand-eye calibration.}
\label{fig:I2IT_example}
\end{figure}

Prior to performing DRHeC, the captured images must be transformed into the simulation-image domain, as shown in Fig.~\ref{fig:differentiable rendering}.
To train the I2IT network, we utilized a dataset comprising 10 robot configurations, each associated with 147 camera poses, resulting in a total of 1,470 images.
Robot masks were extracted from the observed images, and these RGB masks were paired with their corresponding simulation images to train the I2IT model.
While paired datasets are not strictly required for I2IT training, empirical evidence suggests that paired datasets often lead to improved performance.
To refine the training dataset, a preprocessing step was introduced, as the binary masks share the same mask area as the RGB masks.
In this process, a preliminary binary-mask-based calibration was used only to improve the alignment between the observed images and the rendered simulation images when constructing the I2IT training pairs, thereby ensuring more accurate color correspondence.
This preprocessing calibration was not used in the subsequent DRHeC optimization.
The real-to-simulation paired dataset collected and annotated from the UR5e experiments is used as one dataset in this work, while the additional DENSO dataset described in Section VI.E serves as another real-world dataset for I2IT training and evaluation.

The output of the I2IT network is shown in Fig.~\ref{fig:I2IT_example}.
All 10 robot configurations were successfully transformed from the real-image domain to the simulation-image domain, preserving geometric information while altering only color correspondences.
There were no changes to the poses, apart from minor discrepancies resulting from mask extraction and slight differences in the model configurations between the real and simulation robots.

\begin{figure*}[ht]
    \centering
    \captionsetup[subfloat]{labelformat=empty}

    \begin{tabular}{c @{\hspace{4mm}} *{5}{c@{\hspace{2mm}}}}
        \rotatebox{90}{\textbf{Pose 1}} &
        \subfloat[]{\includegraphics[width=0.18\linewidth]{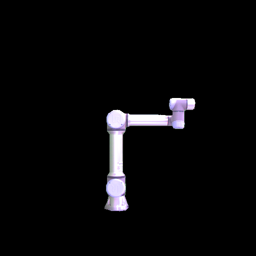}} &
        \subfloat[]{\includegraphics[width=0.18\linewidth]{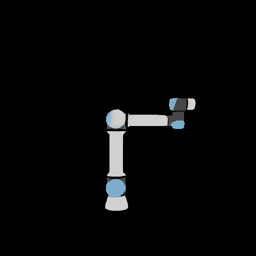}} &
        \subfloat[]{\includegraphics[width=0.18\linewidth]{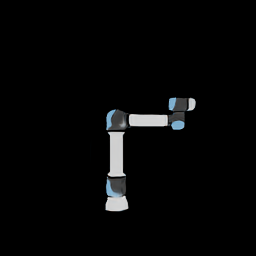}} &
        \subfloat[]{\includegraphics[width=0.18\linewidth]{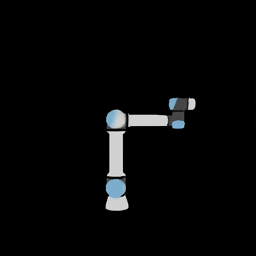}} &
        \subfloat[]{\includegraphics[width=0.18\linewidth]{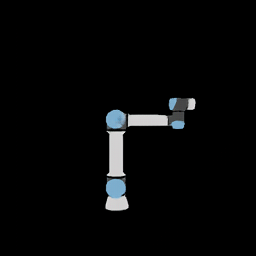}} \\
        \rotatebox{90}{\textbf{Pose 2}} &
        \subfloat[RGB Mask]{\includegraphics[width=0.18\linewidth]{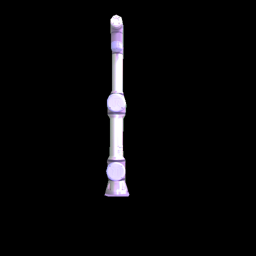}} &
        \subfloat[CycleGAN]{\includegraphics[width=0.18\linewidth]{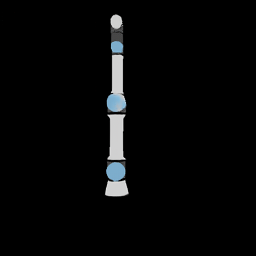}} &
        \subfloat[CUT]{\includegraphics[width=0.18\linewidth]{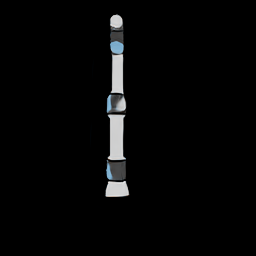}} &
        \subfloat[GeoMaskGAN]{\includegraphics[width=0.18\linewidth]{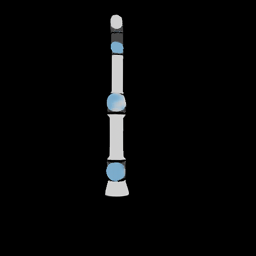}} &
        \subfloat[DRHeC]{\includegraphics[width=0.18\linewidth]{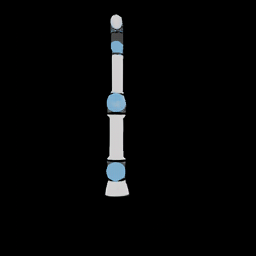}} \\
    \end{tabular}

    \caption{
        Visual comparison of real-world RGB images and predicted simulation-like images from different I2IT methods across robot poses.
        DRHeC shows better performance in color consistency for challenging robot configurations compared to other methods.
    }
    \label{fig:I2IT_result_comparision}
\end{figure*}

\begin{table}[htbp]
\caption{Performance of I2IT methods in the real-world experiment.}
\label{table:I2IT_result}
\centering
\begin{tabularx}{\linewidth}{l *{5}{>{\centering\arraybackslash}X}}
\hline
Method & MSE & MAE & IoU & PSNR (dB) & SSIM \\
\hline
CycleGAN \cite{zhu2017unpaired} & 294.35 & \textbf{3.19} & 0.9300 & 23.68 & \textbf{0.9725} \\
CUT \cite{park2020contrastive} & 354.60 & 3.46 & 0.9284 & 22.76 & 0.9680 \\
GeoMaskGAN \cite{lu2022geometry} & 304.74 & 3.26 & 0.9286 & 23.54 & 0.9705 \\
DRHeC (ours) & \textbf{292.12} & \textbf{3.19} & \textbf{0.9350} & \textbf{23.70} & 0.9722 \\
\hline
\end{tabularx}
\end{table}

As shown in Table~\ref{table:I2IT_result}, the proposed DRHeC method outperforms the state-of-the-art I2IT approaches across most evaluation metrics. While its SSIM score reaches 0.9722, slightly lower than that of CycleGAN at 0.9725, DRHeC achieves the lowest MSE and MAE, the highest PSNR, and a notably higher mask IoU, indicating superior geometric consistency.
In addition to these quantitative results, the visual comparisons in Fig.~\ref{fig:I2IT_result_comparision} further demonstrate that DRHeC more accurately preserves both the appearance fidelity and geometric structure of simulation-style images. Notably, our method maintains consistent color translation even under challenging robot poses, while other methods often fail around joint regions.
This robust I2IT performance provides a reliable image translation backbone and facilitates accurate downstream RGB-based differentiable renderer hand-eye calibration.

\subsection{Grasping Experiments}
\label{subsec:Grasping Experiments}
Grasping tasks are essential in many robotic applications, where accurate hand-eye calibration is crucial for enabling the robot to effectively pick and place objects. The accuracy of the calibration method directly impacts the robot's ability to estimate the object's pose and perform a successful grasp. Therefore, the grasping experiments were designed to evaluate the performance of different calibration methods in guiding the robot to grasp objects accurately.

For the differentiable rendering-based method, we employed two types of robot configurations, three marker positions, and nine camera poses, resulting in a total of 54 combinations. In comparison, the marker-based method uses images captured from 10 different robot poses, leading to 18 trials. Three marker positions were used in each trial, yielding another 54 combinations, allowing for a comparison with the differentiable rendering-based approach. The results of the grasping experiments are summarized in Table~\ref{table:exp_ur5e}.

The experimental results show that DRHeC substantially improves over the differentiable renderer hand-eye calibration baseline EasyHeC.
Compared with the closed-form marker-based baselines included in this setup, DRHeC also shows strong performance, while Single3D (iterations) still achieves the highest success rate when fiducial markers are available.
Under our experimental setup, when using 10 images for calibration, the traditional marker-based methods (Tsai et~al. and Daniilidis et~al.) achieve success rates of 5.6\% and 9.3\%, respectively.
The recent closed-form marker-based methods, Single3D (closed-form) and Wang et~al., also exhibit low success rates, achieving only 3.7\% and 24.1\% success rates, respectively.
In contrast, the iterative variant Single3D (iterations) performs substantially better and achieves the highest success rate among all methods.
The relatively low success rates of some marker-based baselines may be associated with the specific imaging conditions in our setup, particularly the approximately 2 m distance between the camera and the robot base.
At this distance, the fiducial markers occupy relatively small image regions, which may reduce the accuracy of marker detection and pose estimation.
Moreover, their performance may also be affected by measurement-pose selection and marker precision in our experimental setup.
The strong performance of Single3D (iterations) suggests that its iterative refinement is beneficial under the tested conditions.

In contrast, the differentiable rendering method does not rely on fiducial markers during hand-eye calibration.
Instead, it leverages the robot's own geometry for hand-eye calibration, where pose estimation is determined using the entire robot mask rather than relying on specific corner points or feature points.
While the EasyHeC method performs better with a success rate of 42.6\%, it still falls short of optimal performance due to the limitations of binary-mask-based supervision, which provides limited geometric information and lacks discriminative appearance cues.
In comparison, DRHeC achieves a success rate of 88.9\%, substantially improving over EasyHeC under this experimental setup.
It also outperforms several closed-form marker-based baselines, although Single3D (iterations) remains the strongest overall method in these grasping trials.
These results support the effectiveness of the proposed markerless hand-eye calibration under the tested conditions.
Although Single3D (iterations) attains the highest overall success rate, it relies on fiducial-marker observations.
In contrast, DRHeC achieves strong performance without relying on fiducial markers for solving the hand-eye calibration problem, although AprilTags are used only for downstream object-pose estimation in the task evaluation setup.
This demonstrates the effectiveness of the proposed differentiable-renderer hand-eye calibration framework.

\begin{table*}[ht]
\centering
\caption{Success rates of different experiments on UR5e robot.}
\label{table:exp_ur5e}
\resizebox{\textwidth}{!}{
\begin{tabular}{c|ccccc|cc}
\hline
\multirow{2}{*}{Method}
& \multicolumn{5}{c|}{Marker-based}
& \multicolumn{2}{c}{Differentiable-rendering-based (Markerless)} \\
\cline{2-8}
& Tsai et~al.~\cite{tsai1989new}
& Daniilidis et~al.~\cite{daniilidis1999hand}
& Single3D (closed-form)~\cite{jin2024hand}
& Single3D (iterations)~\cite{jin2024hand}
& Wang et~al.~\cite{wang2024dual}
& EasyHeC~\cite{chen2023easyhec}
& DRHeC \\
\hline
Grasping & 5.6\% & 9.3\% & 3.7\% & 96.3\% & 24.1\% & 42.6\% & 88.9\% \\
Insertion & 5.6\% & 9.3\% & 1.8\% & 81.5\% & 16.7\% & 9.3\% & 57.4\% \\
\hline
\end{tabular}
}
\end{table*}

\subsection{Insertion Experiments}
\label{subsec:Insertion Experiments}

Insertion tasks are also critical for robot manipulators, as they require precise alignment and positioning to ensure successful interaction with objects.
Such tasks are common in manufacturing, assembly lines, and maintenance operations, where robots are required to insert components into specific slots or holes with high accuracy.
The ability to perform these tasks efficiently and precisely is a key indicator of a robot's manipulation capabilities, especially when dealing with complex geometries and tight tolerances.
In this set of experiments, we assessed the capability of each method to achieve accurate and successful insertion.
The setup for the insertion experiment follows the same configuration as the grasping task.

As shown in Table~\ref{table:exp_ur5e}, the success rates for the insertion experiment follow a similar trend to those observed in the grasping task.
The marker-based methods (Tsai et~al., Daniilidis et~al., Single3D (closed-form), Wang et~al.) again show low success rates of 5.6\%, 9.3\%, 1.8\% and 16.7\%, respectively.
Single3D (iterations) again maintains the highest success rate, consistent with its performance in the grasping task.
The EasyHeC method also performs poorly, with a success rate of 9.3\%.
In the insertion task, DRHeC achieves 57.4\%, showing a substantial improvement over EasyHeC and higher success rates than several closed-form marker-based baselines under the tested setup, while remaining below Single3D (iterations).
The lower success rate compared to the grasping task is due to the increased difficulty of the insertion task, which requires more precise alignment.
Additionally, we apply a stricter criterion for success in insertion: any collision with the object is considered a failure.
These factors contribute to the lower overall success rate in the insertion experiment.

Although both the Daniilidis et~al. method and the EasyHeC method yield the same 9.3\% success rate, they exhibit a notable difference in performance. The EasyHeC method results in only a slight deviation from the target, whereas the marker-based methods, particularly the Tsai and Daniilidis approaches, tend to produce much larger errors, especially in translation.

To better understand the calibration differences among the compared methods, we further conduct a numerical analysis on one representative trial.
Among all methods, Single3D (iterations) yields the most accurate and stable calibration result under the tested setup, and its solution is therefore used only as a reference calibration result for relative comparison in this representative trial. It should not be interpreted as absolute ground truth.
The DRHeC method shows a translation error of $0.03$ m, and the EasyHeC method shows a translation error of $0.05$ m, both smaller than those of the Tsai et~al. method ($0.14$ m), the Daniilidis et~al. method ($0.16$ m), the Wang et~al. method ($0.10$ m), and the Single3D (closed-form) method ($0.52$ m).
Rotation errors show a similar trend: DRHeC and EasyHeC yield smaller deviations, whereas other closed-form methods exhibit significantly larger angular errors.
(with rotation errors of $0.02$, $0.03$, $0.13$, $0.09$, $0.04$, and $0.85$ rad corresponding to DRHeC, EasyHeC, Tsai, Daniilidis, Wang, and Single3D (closed-form), respectively).
Using Single3D (iterations) as the reference, these results indicate that DRHeC provides the closest estimation in this trial, followed by EasyHeC, while the compared closed-form marker-based methods show larger deviations overall.
The compared closed-form marker-based methods also exhibit larger translation and rotation errors, which likely contributed to their insertion failures.
Although EasyHeC still does not achieve successful insertion, it shows a smaller deviation from the reference than several closed-form marker-based methods in this representative trial.

In conclusion, DRHeC shows strong performance among the evaluated differentiable-rendering methods and achieves higher success rates than several closed-form marker-based methods under the tested setup, while remaining below Single3D (iterations).
This supports the effectiveness of DRHeC for successful insertion without relying on fiducial markers.

\begin{figure}[t]
  \captionsetup[subfloat]{labelformat=empty}
  \centering
  \subfloat[(1-a)]{
    \includegraphics[width=0.49\linewidth]{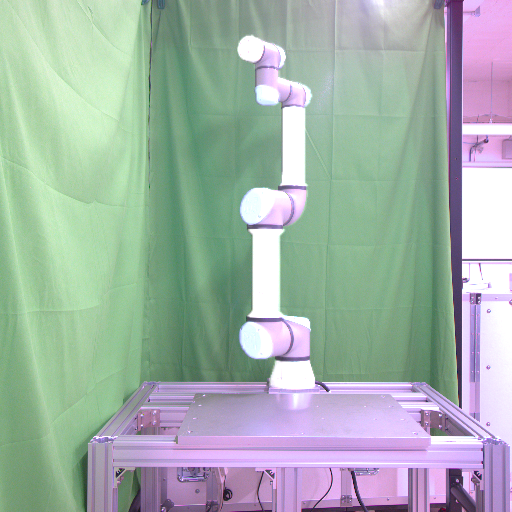}
    \label{fig:reprojection_DRHeC1}
  }
  \subfloat[(1-b)]{
    \includegraphics[width=0.49\linewidth]{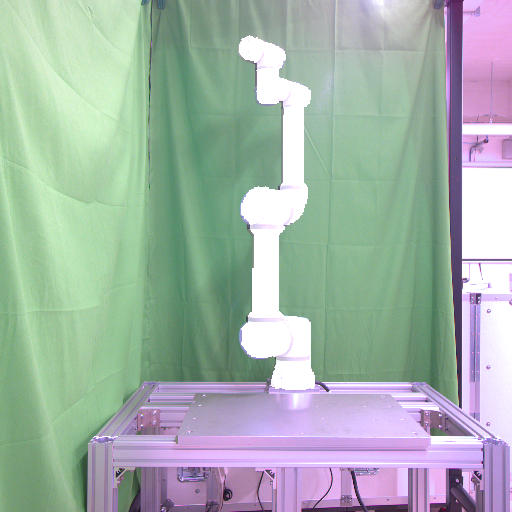}
    \label{fig:reprojection_easyhec1}
  } \\[0.5ex]  %

  \subfloat[(2-a)]{
    \includegraphics[width=0.49\linewidth]{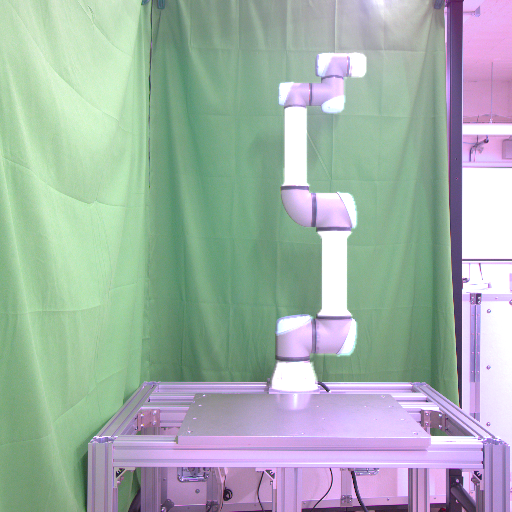}
    \label{fig:reprojection_DRHeC2}
  }
  \subfloat[(2-b)]{
    \includegraphics[width=0.49\linewidth]{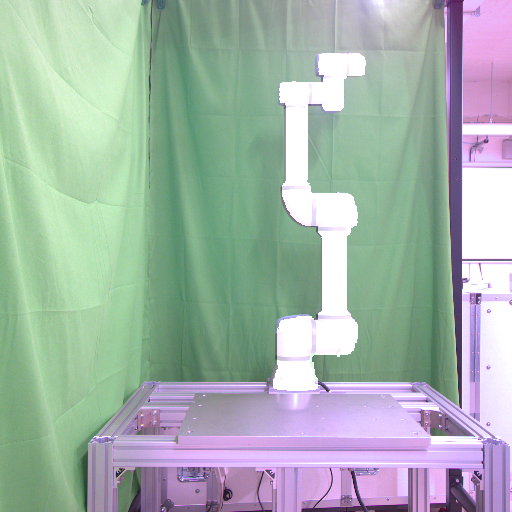}
    \label{fig:reprojection_easyhec2}
  }

  \caption{Reprojection of calibration results on real-world images for two scenarios (1 and 2). Results from DRHeC are shown in the left column (1-a and 2-a), whereas those from EasyHeC are shown in the right column (1-b and 2-b).}
  \label{fig:reprojection}
\end{figure}

\subsection{DENSO VS060 Experiments}
\label{subsec:Denso Robot Experiments}

To provide an additional validation beyond the UR5e experiments, we conducted 18 parallel trials using the DENSO VS060 robot (DENSO WAVE Inc., Aichi, Japan) equipped with an OnRobot RG2 gripper (OnRobot A/S, Odense, Denmark).
In contrast to the UR5e robot, which is visually distinctive due to richer color variation, the VS060 has a nearly uniform appearance across all joints, with only minor visual differences at the end effector.
This reduced color variation makes RGB-based optimization more challenging and therefore provides a useful additional test case for the proposed framework.
The success rates for each method are presented in Table~\ref{table:exp_vs060}.

\begin{table*}[ht]
\centering
\caption{Success rates of different experiments on DENSO VS060 robot.}
\label{table:exp_vs060}
\resizebox{\textwidth}{!}{
\begin{tabular}{c|ccccc|cc}
\hline
\multirow{2}{*}{Method}
& \multicolumn{5}{c|}{Marker-based}
& \multicolumn{2}{c}{Differentiable-rendering-based ( Markerless )} \\
\cline{2-8}
& Tsai et~al.\cite{tsai1989new} & Daniilidis et~al.\cite{daniilidis1999hand}
& Single3D (closed-form)\cite{jin2024hand}
& Single3D (iterations)\cite{jin2024hand}
& Wang et~al.\cite{wang2024dual}
& EasyHeC\cite{chen2023easyhec} & DRHeC \\
\hline
Grasping & 11.1\% & 11.1\% & 38.9\% & 94.4\% & 44.4\% & 55.6\% & 77.8\% \\
Insertion & 11.1\% & 11.1\% & 27.8\% & 83.3\% & 22.2\% & 33.3\% & 77.8\% \\
\hline
\end{tabular}
}
\end{table*}

Compared with the baseline methods, DRHeC achieved strong success rates in both grasping and insertion tasks, outperforming EasyHeC by 22.2 and 44.5 percentage points, respectively, although Single3D (iterations) achieved the highest success rates overall.
This demonstrates the robustness of DRHeC across different robotic platforms.
Despite the lack of distinctive visual features on the DENSO VS060, DRHeC still performed strongly, highlighting its ability to handle challenging visual conditions and generalize effectively.
We also provide a relative quantitative comparison using one representative trial on the DENSO VS060 robot, taking the Single3D (iterations) result only as a reference calibration result under the tested setup, rather than as absolute ground truth.
The translation errors for DRHeC, EasyHeC, Tsai et~al., Daniilidis et~al., Wang et~al., and Single3D (closed-form) are
$0.03$, $0.05$, $0.20$, $0.08$, $0.07$, and $0.05$ m, respectively.
The corresponding rotation errors are $0.03$, $0.04$, $0.03$, $0.01$, $0.01$, and $0.09$ rad.
Considering both translation and rotation, DRHeC provides the closest alignment to the Single3D (iterations) reference result, followed by EasyHeC, while the compared closed-form marker-based methods tend to show larger deviations.
Overall, these results support the practicality of DRHeC for offline or setup-time industrial hand-eye calibration scenarios.

\subsection{Discussion}
\label{subsec:Discussion}

The experimental results show that DRHeC substantially improves over EasyHeC under the present setup and achieves higher success rates than several closed-form marker-based baselines under the tested conditions.
This improvement is attributed to DRHeC's ability to integrate color and mask geometric features.
Marker-based methods, while computationally efficient, may be sensitive to marker occlusion and variations in imaging conditions.
Moreover, in our experimental setup, the relatively long calibration distance (2 meters), limited camera poses (only 10), and the use of printed markers with limited precision may have further degraded accuracy and constrained the performance of traditional marker-based methods in this study.
Similarly, EasyHeC achieves higher success rates than several closed-form marker-based baselines under this experimental setup, but it still suffers from several limitations, including restricted supervision from binary mask configurations and the lack of color information, which can reduce robustness and precision.
In contrast, DRHeC improves calibration performance by incorporating RGB cues and mask geometric features.

While the proposed DRHeC method achieves strong performance, there are still areas for improvement.
As shown in Fig.~\ref{fig:reprojection}, DRHeC exhibits slight misalignment between the robot in the simulation and the real world.
This issue arises from the limitations of the I2IT method, which struggles to fully transform the real-world robot into a simulation-like image.
Adding more diverse input to the I2IT training set may help address the misalignment issues by better capturing real-world variations.
Despite CycleGAN's ability to train I2IT transformations without paired datasets, discrepancies still arise in the simulation robot model, such as differences in color configuration and texture that cannot be fully avoided.
Additionally, imperfect mask extraction introduces slight errors in color correspondence.
Optimization remains challenging as it may fall into local minima when using only a single image.

From the perspective of single-frame observability, the calibration is more likely to be well constrained when perturbations of \({}^{c}T_b\) produce sufficiently distinguishable changes in the rendered RGB and mask observations under a known robot configuration.
In practice, this is more likely to hold when the robot structure is sufficiently visible, self-occlusion is limited, projection symmetry is weak, and the image contains informative RGB and mask gradients.
In contrast, degradation may occur when the known robot configuration and camera viewpoint produce highly similar observations under different perturbations of \({}^{c}T_b\), for example under projection symmetry, severe self-occlusion, limited visible structure, or weak appearance variation.
In this context, pose ambiguity should be understood only as one possible qualitative manifestation of degraded single-frame observability.
A formal observability analysis and a quantitative tolerance boundary for such degraded cases were beyond the scope of the present study.

\begin{figure}[t]
  \captionsetup[subfloat]{labelformat=empty}
  \centering

  \subfloat[(a)]{
    \includegraphics[width=0.31\linewidth]{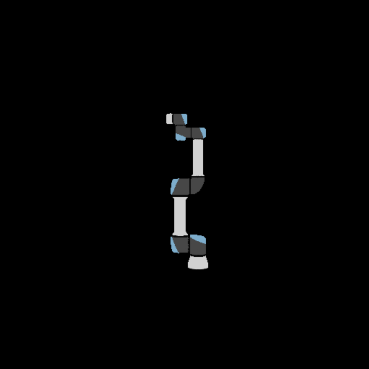}
  }
  \subfloat[(b)]{
    \includegraphics[width=0.31\linewidth]{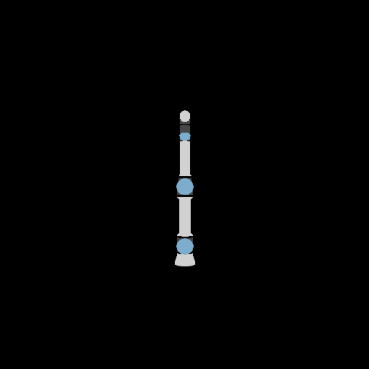}
  }
  \subfloat[(c)]{
    \includegraphics[width=0.31\linewidth]{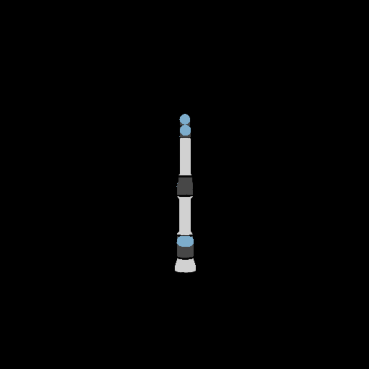}
  }

  \subfloat[(d)]{
    \includegraphics[width=0.31\linewidth]{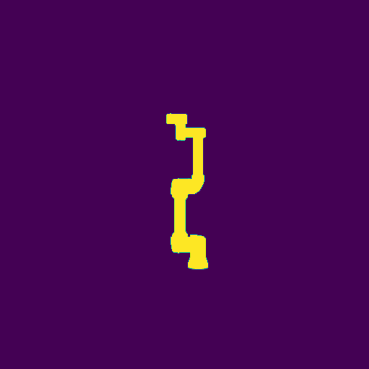}
  }
  \subfloat[(e)]{
    \includegraphics[width=0.31\linewidth]{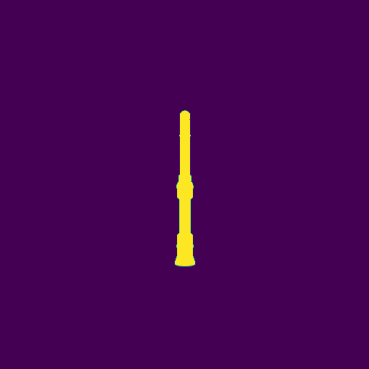}
  }
  \subfloat[(f)]{
    \includegraphics[width=0.31\linewidth]{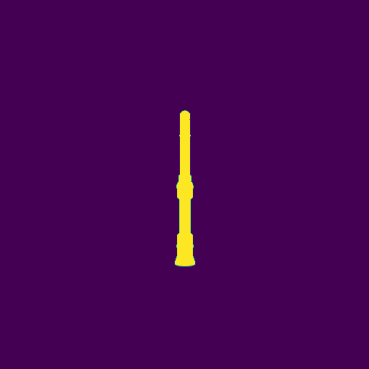}
  }

  \caption{
    Illustrative example of pose ambiguity in binary-mask-based calibration.
    RGB images (a-c) and corresponding binary masks (d-f) represent three distinct robot poses:
    (a,d): [0°, -90°, 0°, -90°, 0°, 0°]; (b,e): [90°, -90°, 0°, -90°, 0°, 0°]; and (c,f): [-90°, -90°, 0°, -90°, 0°, 0°].
    Despite the 180° difference in Joint 1 between (b) and (c), their binary masks (e,f) appear nearly identical due to projection ambiguity.
    Methods relying on binary-mask loss may have difficulty distinguishing these poses during optimization.
    In contrast, DRHeC incorporates RGB information, which can help distinguish such cases.
  }
  \label{fig:ambiguity}
\end{figure}

Pose ambiguity may arise in some binary-mask-based calibration cases.
As illustrated in Fig.~\ref{fig:ambiguity}, distinct robot poses can produce nearly identical binary masks due to projection symmetry and occlusion effects.
The additional color and texture information encoded in the RGB data provides discriminative cues that help distinguish robot poses which are otherwise indistinguishable by binary masks alone.
In such cases, methods relying only on binary-mask supervision may have difficulty distinguishing between different poses, which can reduce calibration accuracy.
In contrast, DRHeC leverages RGB image derivatives to provide additional color and texture cues, which can help distinguish such cases during optimization.
This example illustrates a potential additional benefit of RGB information in cases where binary masks become insufficiently discriminative.
Since this work does not provide a systematic quantitative evaluation specifically targeting such ambiguity cases, we present this observation as a qualitative example rather than a central validated claim.

In addition, although the DENSO VS060 robot provides minimal color variation, DRHeC still achieved strong calibration performance.
This result suggests that DRHeC effectively leverages mask geometric features, specifically the mask centroid and mask area, to provide global information for gradient-based optimization, even when explicit color information is limited.
This capability is particularly advantageous in industrial scenarios where robots may have uniform color schemes.
Furthermore, in practical applications, robots often have manufacturer labels or markings, and end-effectors may be equipped with tools that introduce additional color variations.
In the current pipeline, the influence of such appearance changes mainly depends on the I2IT stage.
If these changes do not significantly affect feature extraction or the generation of real-to-sim images, then they are not expected to have a substantial impact on the subsequent DRHeC calibration.
However, if they affect the I2IT translation process, then the generated sim-like images may no longer satisfy the assumptions of the current experimental setting, and the final calibration performance may also be affected.
A dedicated evaluation with deliberate appearance modifications was beyond the scope of the present study, and a more systematic evaluation under deliberate color and texture variations will be investigated in future work.
Although the current experiments on Baxter, UR5e, and DENSO VS060 provide initial evidence of applicability across different robot settings, the generalization capability of the framework has not yet been systematically evaluated on a broader range of robot platforms.
More extensive validation on a broader range of robot platforms will be an important direction for future work.

Our method assumes reasonably stable indoor illumination.
Although it is not designed to handle illumination variations, it remains stable under the mild indoor lighting fluctuations present during all our real-world experiments.
However, extreme lighting conditions or occlusions on the robot mask may affect the reliability of the pipeline.
When the scene becomes excessively bright or dark, the observed images may contain more noise and reduced contrast, which will negatively influence both the segmentation results and the I2IT translation process.
Meanwhile, occlusions or blocking on the robot mask will interfere with the mask consistency used by I2IT, causing the translated sim-like images to contain noticeable errors.
These inaccurate translated images may then propagate through the differentiable rendering pipeline and ultimately reduce the accuracy of the final hand-eye calibration result.

We further evaluated the sensitivity of DRHeC to geometric model mismatch in simulation by rendering reference images with globally anisotropic Gaussian scaling perturbations with standard deviations of 0\%, 1\%, 3\%, and 5\%.
The experiment comprised 20 paired trials using the UR5e robot model in the simulation environment.
The mean translation errors were 7.41, 21.92, 53.05, and 86.13 mm, respectively, while the corresponding mean rotation errors were 0.0197, 0.0190, 0.0287, and 0.0396 rad.
These results indicate that a 1\% geometric mismatch already substantially degrades translation accuracy, whereas the effect on rotation accuracy appears to become more noticeable at larger mismatch levels.
A systematic quantitative evaluation of the remaining factors, including controlled illumination changes, external occlusions, segmentation failures, and deliberate robot appearance variations, remains an important direction for future work.

In terms of computational complexity, our differentiable rendering-based optimization requires approximately 7.19 seconds per 1000 iterations for DRHeC, which is comparable to EasyHeC (7.18 seconds) and significantly faster than IPE (78.25 seconds).
Under the 2000-iteration setting used in this study, DRHeC requires about 14 seconds per calibration.
Since this cost is incurred only once during calibration and the estimated hand-eye transformation can be reused for subsequent manipulation tasks, we consider this runtime practical for offline hand-eye calibration or occasional recalibration.
Further acceleration would be beneficial for online recalibration or fully automated high-throughput deployment.

\section{Ablation Study}

\begin{figure}[t]
  \captionsetup[subfloat]{labelformat=empty}
  \centering

  \subfloat[]{\includegraphics[width=0.48\linewidth]{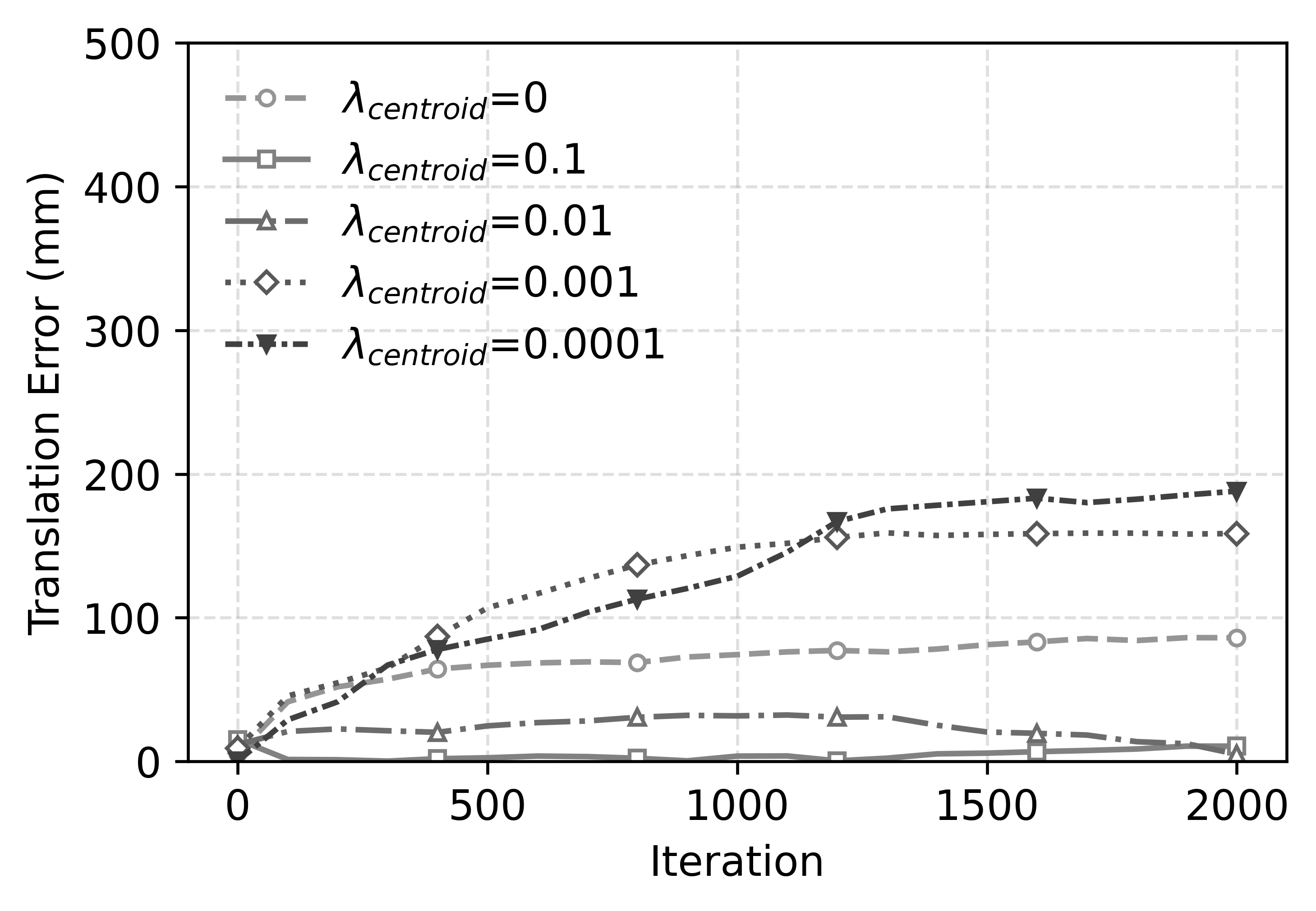}}
  \subfloat[]{\includegraphics[width=0.48\linewidth]{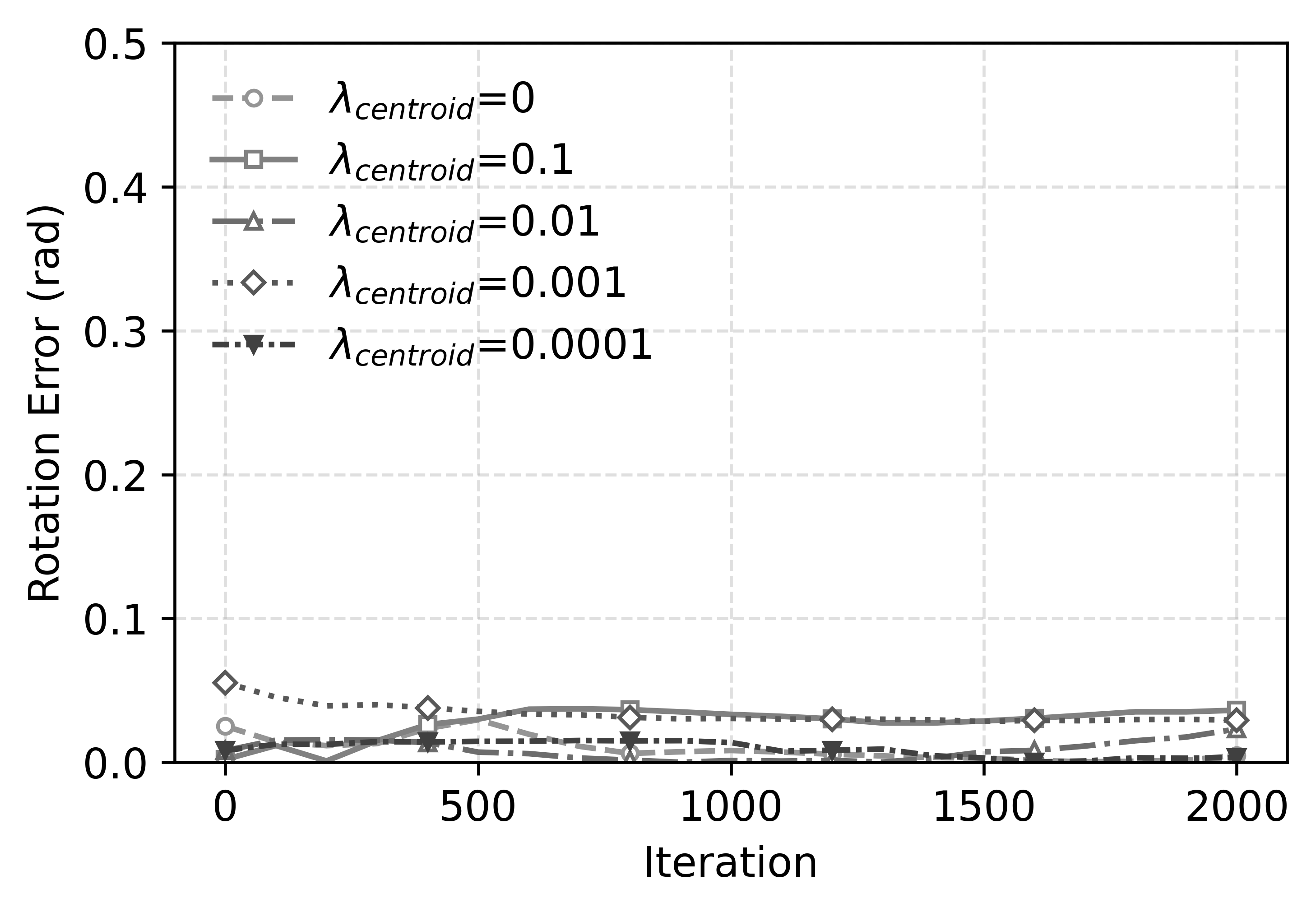}}\\[-1.5ex]
  \makebox[\linewidth]{(a)}
  \subfloat[]{\includegraphics[width=0.48\linewidth]{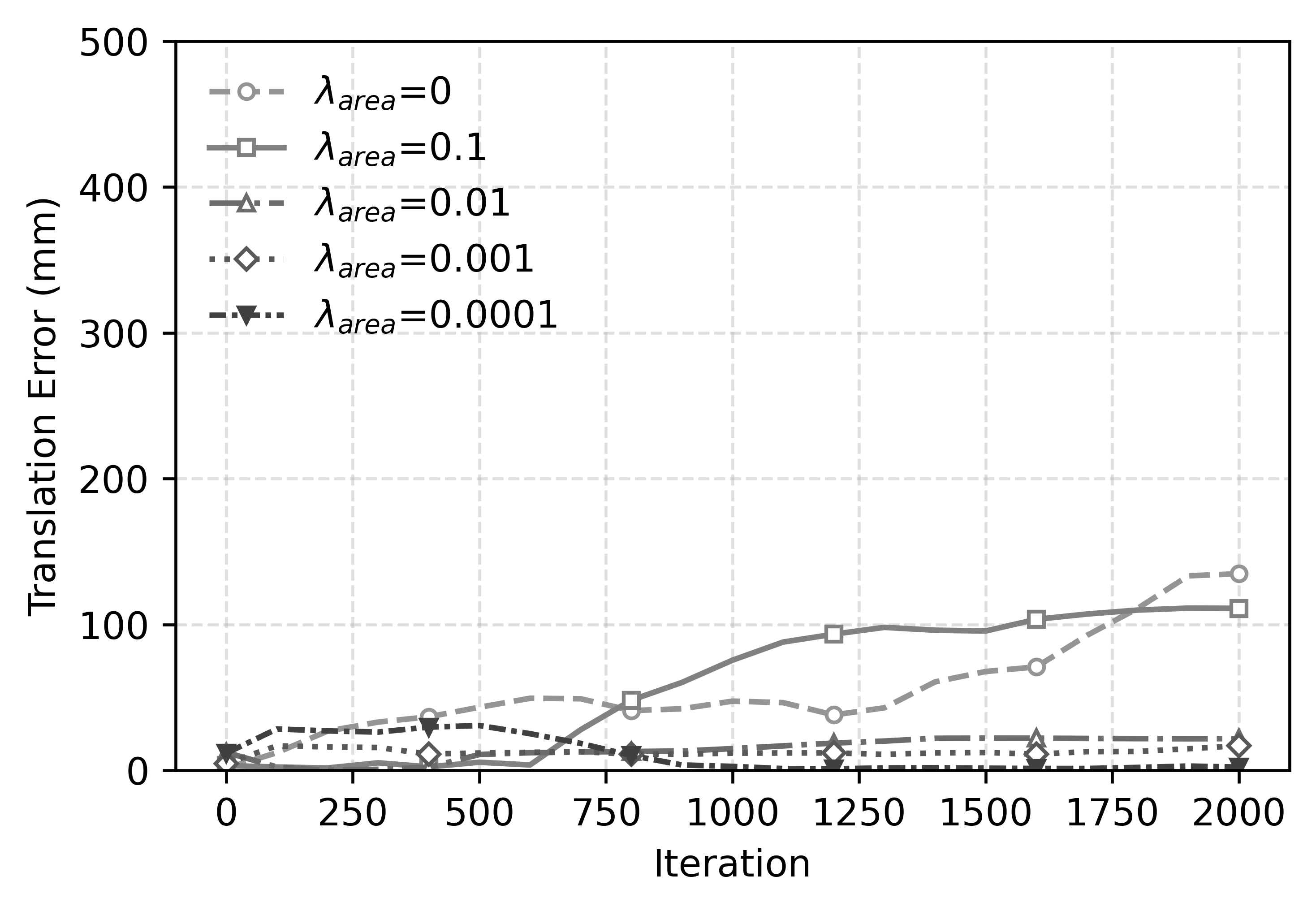}}
  \subfloat[]{\includegraphics[width=0.48\linewidth]{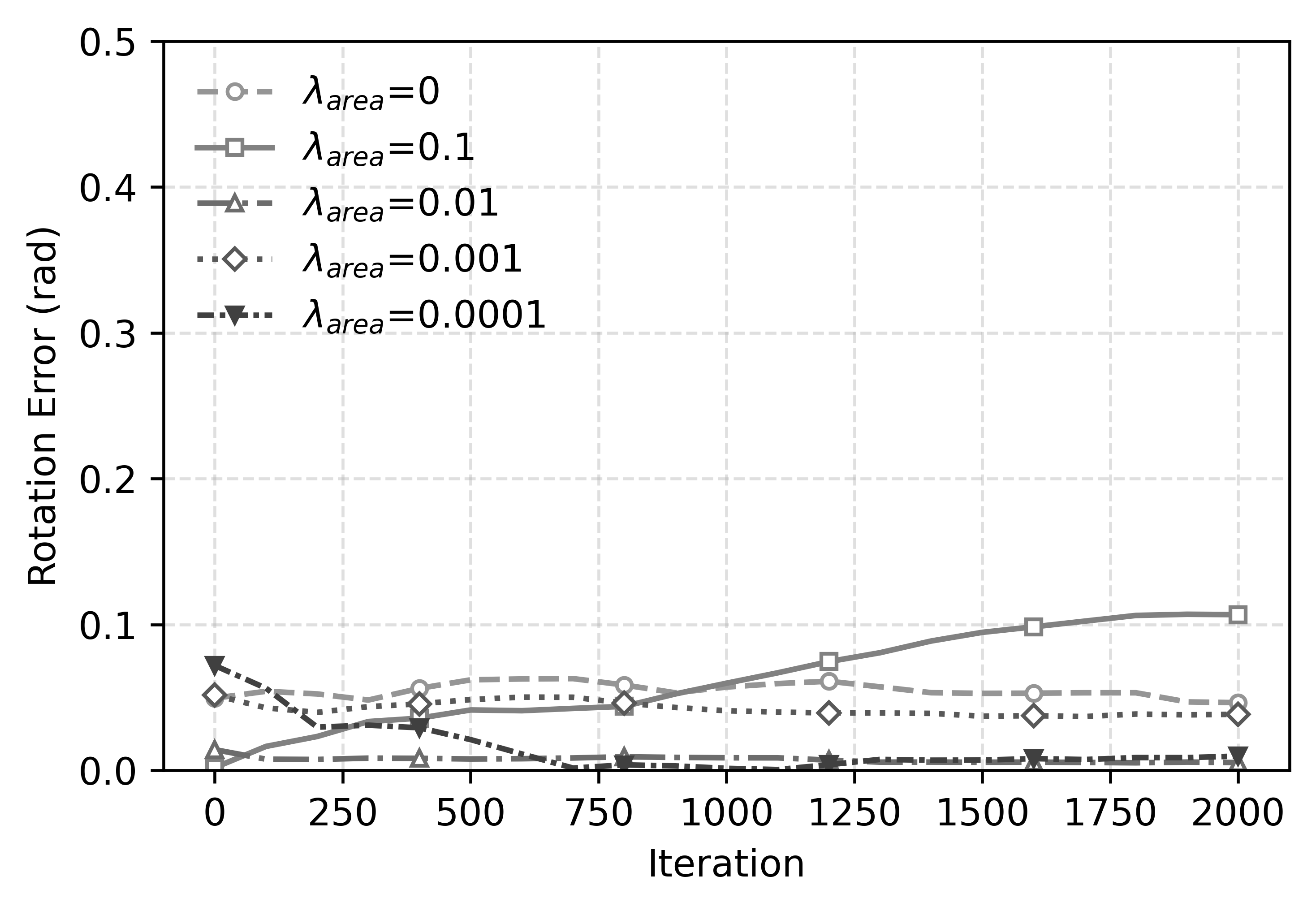}}\\[-1.5ex]
  \makebox[\linewidth]{(b)}
  \subfloat[]{\includegraphics[width=0.48\linewidth]{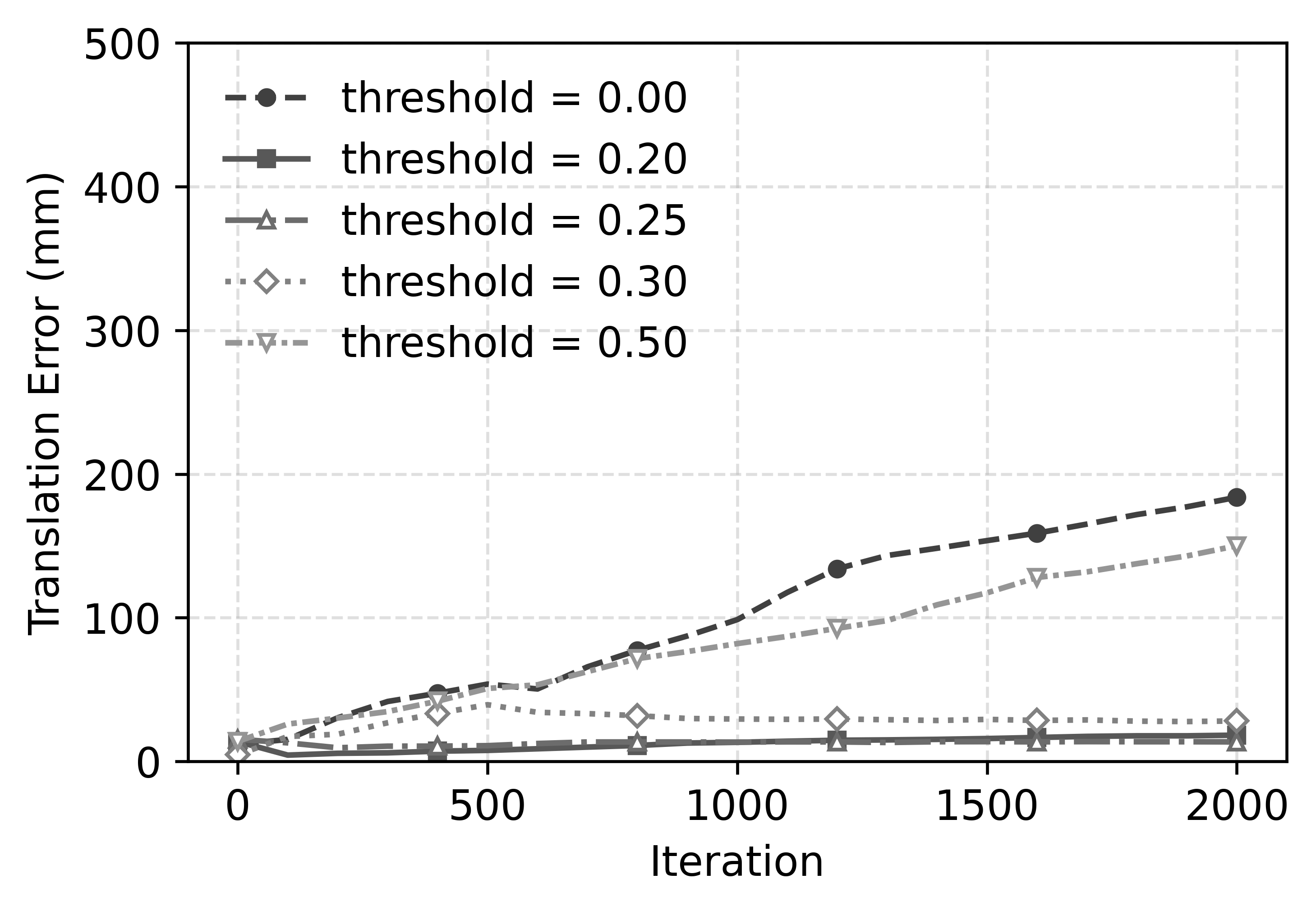}}
  \subfloat[]{\includegraphics[width=0.48\linewidth]{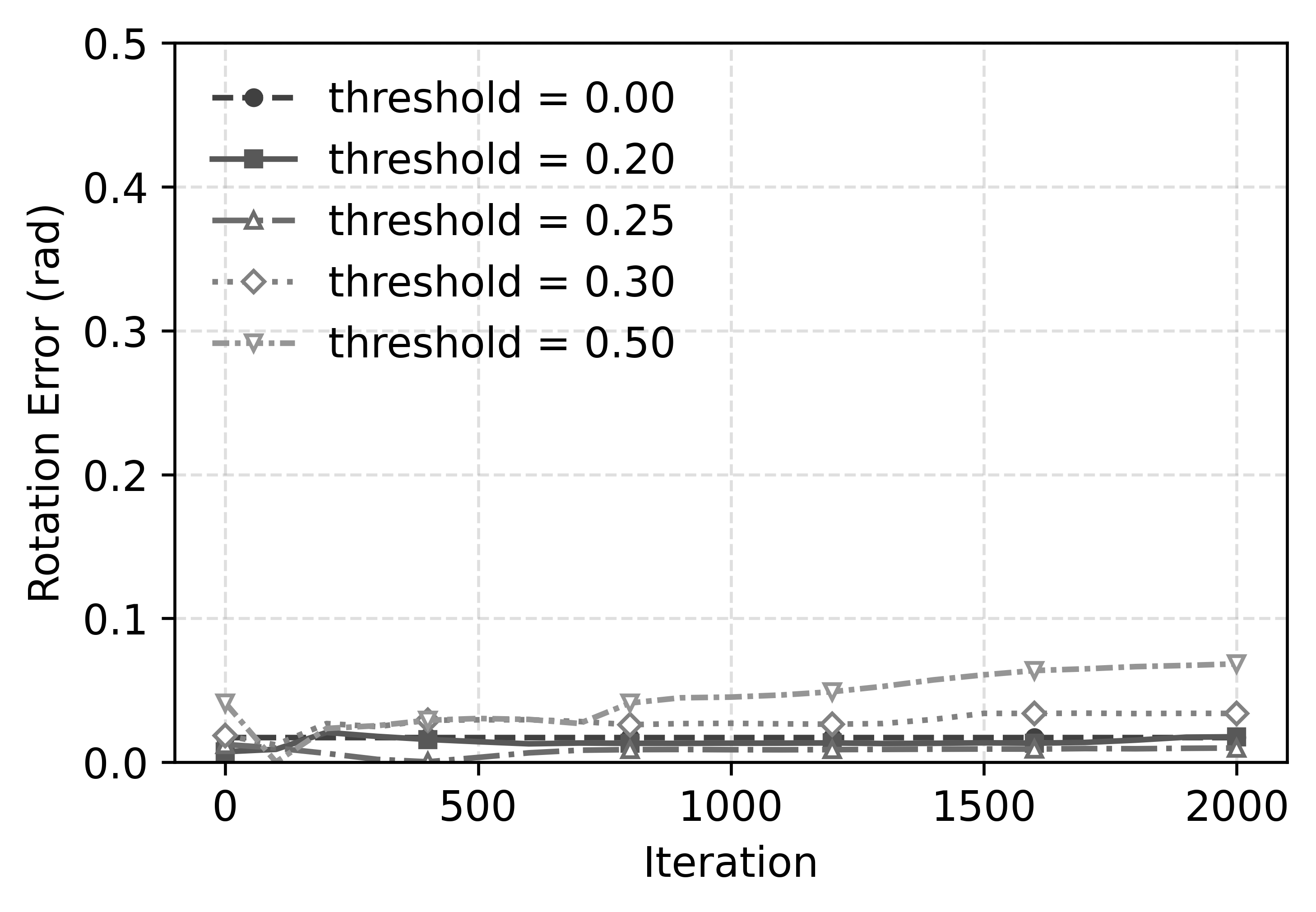}}\\[-1.5ex]
  \makebox[\linewidth]{(c)}

  \caption{Ablation study on (a) $\lambda_{\text{centroid}}$, (b) $\lambda_{\text{area}}$, and (c) threshold for the two-step optimization parameters.
  Each pair shows translation and rotation errors, respectively.}
  \label{fig:ablation_study}
\end{figure}

The simulation experiments in Sec.~\ref{subsec:Synthetic Dataset} already form an ablation study over different combinations of loss terms.
Specifically, we compare four configurations:
(1) EasyHeC (binary mask only),
(2) RGB (using RGB mask without geometric terms),
(3) RGB + mask centroid, and
(4) our full DRHeC method (RGB + mask centroid + mask area).
The results consistently show that DRHeC achieves the best performance, followed by RGB + mask centroid and RGB alone.
This performance hierarchy clearly demonstrates the importance of incorporating mask geometric terms for improving stability and accuracy.

We also conduct an ablation study to determine the appropriate hyperparameters for the loss function in Eq.~\ref{eq:loss_function} under the LP and MP settings, as the HP setting produces errors that are unrealistically large for practical hand-eye calibration applications.
We set $\lambda_{\mathrm{rgb}}=1$ as the reference weight.
Since the RGB, centroid, and area losses are defined over different quantities and therefore have different numerical scales, logarithmically spaced candidate weights were evaluated for $\lambda_{\mathrm{centroid}}$ and $\lambda_{\mathrm{area}}$.
The final values were subsequently selected based on the ablation results.
All ablation experiments are implemented using our DRHeC framework, and each configuration is evaluated over 20 independent parallel trials to ensure statistical reliability.

As shown in Table~\ref{table:ablation_lambda_centroid}, Table~\ref{table:ablation_lambda_area} and Table~\ref{table:ablation_threshold}, the proposed hyperparameter configuration achieves the most robust and accurate calibration results. Specifically, the parameters $\lambda_{\text{centroid}} = 0.01$, $\lambda_{\text{area}} = 0.0001$, and threshold $=0.25$ yield the lowest translation errors, while maintaining comparable rotation accuracy across all conditions.
Compared with the baseline configuration without these loss terms, our configuration significantly improves translation accuracy. When $\lambda_{\text{centroid}} = 0$, the translation error increases from 5.05 mm to 86.06 mm, indicating that the mask centroid loss term is essential for providing a geometric constraint that stabilizes convergence.
Likewise, when $\lambda_{\text{area}} = 0$, the translation error increases from 2.30 mm to 134.86 mm, demonstrating that the mask area loss term contributes to scale alignment between the rendered and observed masks.
For the threshold parameter, the results show that the proposed two-step optimization performs well over a reasonable range of threshold values.
In particular, threshold $=0.20$, $0.25$, and $0.30$ all achieve relatively low errors, while threshold $=0.25$ gives the best overall result.
This indicates that the method is not overly sensitive to the specific threshold value.
Consistent with the optimization behavior discussed in Sec.~IV.B, excessively small or large thresholds may degrade performance by delaying rotation refinement or activating rotation updates too early.
Therefore, the main advantage lies in the proposed coarse-to-fine two-step optimization strategy, while $0.25$ is regarded as a practical choice within a stable operating range rather than as a theoretically unique value.
In addition, the same hyperparameter configuration was used for all simulation and real-world experiments without environment-specific retuning.
The experimental results indicate that the selected parameters remain effective across different robot models and in both simulation and real-world scenarios.

\begin{table}[ht]
\caption{Ablation study results of the $\lambda_{centroid}$ over 2000 iterations.}
\label{table:ablation_lambda_centroid}
\centering
\begin{tabular}{cccccc}
\hline
$\lambda_{centroid}$ & 0 & 0.1 & 0.01 & 0.001 & 0.0001 \\ \hline
Translation & 86.06 & 10.67 & $\mathbf{5.05}$ & 158.62 & 188.10 \\
Rotation & $\mathbf{0.00}$ & 0.04 & 0.02 & 0.03 & $\mathbf{0.00}$ \\
\hline
\end{tabular}
\end{table}

\begin{table}[ht]
\caption{Ablation study results of the $\lambda_{area}$ over 2000 iterations.}
\label{table:ablation_lambda_area}
\centering
\begin{tabular}{cccccc}
\hline
$\lambda_{area}$ & 0 & 0.1 & 0.01 & 0.001 & 0.0001 \\ \hline
Translation & 134.86 & 111.17 & 21.81 & 16.99 & $\mathbf{2.30}$ \\
Rotation & 0.05 & 0.11 & $\mathbf{0.01}$ & 0.04 & $\mathbf{0.01}$ \\
\hline
\end{tabular}
\end{table}

\begin{table}[ht]
\caption{Ablation study results of the threshold for the two-step optimization over 2000 iterations.}
  \label{table:ablation_threshold}
\centering
\begin{tabular}{cccccc}
\hline
Threshold & 0 & 0.20 & 0.25 & 0.30 & 0.50 \\ \hline
Translation (mm) & 183.83 & 18.32 & $\mathbf{13.65}$ & 28.23 & 150.11 \\
Rotation (rad) & 0.02 & 0.02 & $\mathbf{0.01}$ & 0.03 & 0.07 \\
\hline
\end{tabular}
\end{table}

We also perform an ablation study to demonstrate that the I2IT module plays an essential role in the real world experiment, and that without it the calibration results would suffer from large errors.
Since the real world setup does not provide ground truth for the hand-eye transformation, we evaluate the influence of the I2IT module by comparing the rendered images obtained through differentiable calibration using the real captured images and the I2IT transformed images.
A more accurate calibration should produce a rendered image that is closer to the simulated view. As shown in Table~\ref{table:ablation_I2IT}, using I2IT clearly improves all image based metrics, including a large reduction in MSE and MAE as well as notable gains in PSNR and SSIM.
These results indicate that the I2IT module effectively reduces the appearance gap between real and synthetic images, which enables more accurate and stable hand-eye calibration.

\begin{table}[htbp]
\caption{Ablation study results of the I2IT module.}
\label{table:ablation_I2IT}
\centering
\begin{tabularx}{\linewidth}{l *{5}{>{\centering\arraybackslash}X}}
\hline
Method & MSE & MAE & IoU & PSNR (dB) & SSIM \\
\hline
Real world image & 0.0099 & 0.0162 & 0.6111 & 20.88 & 0.9477 \\
I2ITed img & \textbf{0.0026} & \textbf{0.0053} & \textbf{0.8520} & \textbf{25.95} & \textbf{0.9727} \\
\hline
\end{tabularx}
\end{table}

Overall, the ablation study confirms that all loss terms and the I2IT module are necessary, and our chosen configuration achieves the best performance among all candidates.
The selected nonzero weights of the RGB derivative, centroid, and area losses stabilize the optimization process, making the differentiable rendering based hand-eye calibration more robust and less prone to failure.

\section{Conclusion}
\label{sec:Conclusion}
In this paper, we introduce a novel framework for robotic hand-eye calibration that utilizes RGB derivatives and mask geometry features in differentiable rendering, which improves calibration accuracy and optimization stability under binary-mask-based supervision.
Additionally, we propose a modified loss function for training the I2IT network, which preserves essential geometric information.
Experimental results demonstrate that our approach achieves strong accuracy and robustness in both simulations and real-world scenarios.

While the proposed method shows promising results, the reliance on a model in differentiable hand-eye calibration can introduce errors due to model inaccuracies.
To address this, we plan to integrate advanced model reconstruction techniques that will reduce errors and enhance calibration precision.
Additionally, because the camera's field of view often requires capturing the entire robot, our current implementation is limited to the eye-to-hand configuration.
To overcome this limitation, we propose leveraging partial observations of the robot to enable a more practical and flexible eye-in-hand calibration.
This approach would allow for either the reconstruction of the full model or direct calibration using only visible segments, thereby increasing the method's flexibility and applicability in various practical settings.
Another practical consideration is that the current mask preparation and runtime are mainly suitable for offline or setup-time hand-eye calibration rather than online recalibration.
Nevertheless, this setting is consistent with many practical hand-eye calibration scenarios, where calibration is performed before task execution and the estimated transformation is reused afterward.
Future work will focus on automatic mask generation and acceleration of the differentiable-renderer-based hand-eye calibration process.
Finally, the method is mainly applicable to indoor scenes with moderate lighting and unobstructed robot visibility, as extreme illumination or occlusions may require additional preprocessing.
Broader validation beyond the tested settings remains an important direction for future work, including formal observability analysis for arbitrary single-frame configurations, calibration uncertainty evaluation with independent reference measurements, and more comprehensive sensitivity analyses covering broader types of robot-model deviations.
Comparisons with recent model-free or weak-model calibration methods under inaccurate-model conditions are also left for future work.

\bibliographystyle{IEEEtran}
\bibliography{articles.bib}

\makeatletter
\def\@IEEEBIOskipN{10pt}
\makeatother
\begin{IEEEbiography}[{\includegraphics[width=1in,height=1.25in,clip,keepaspectratio]{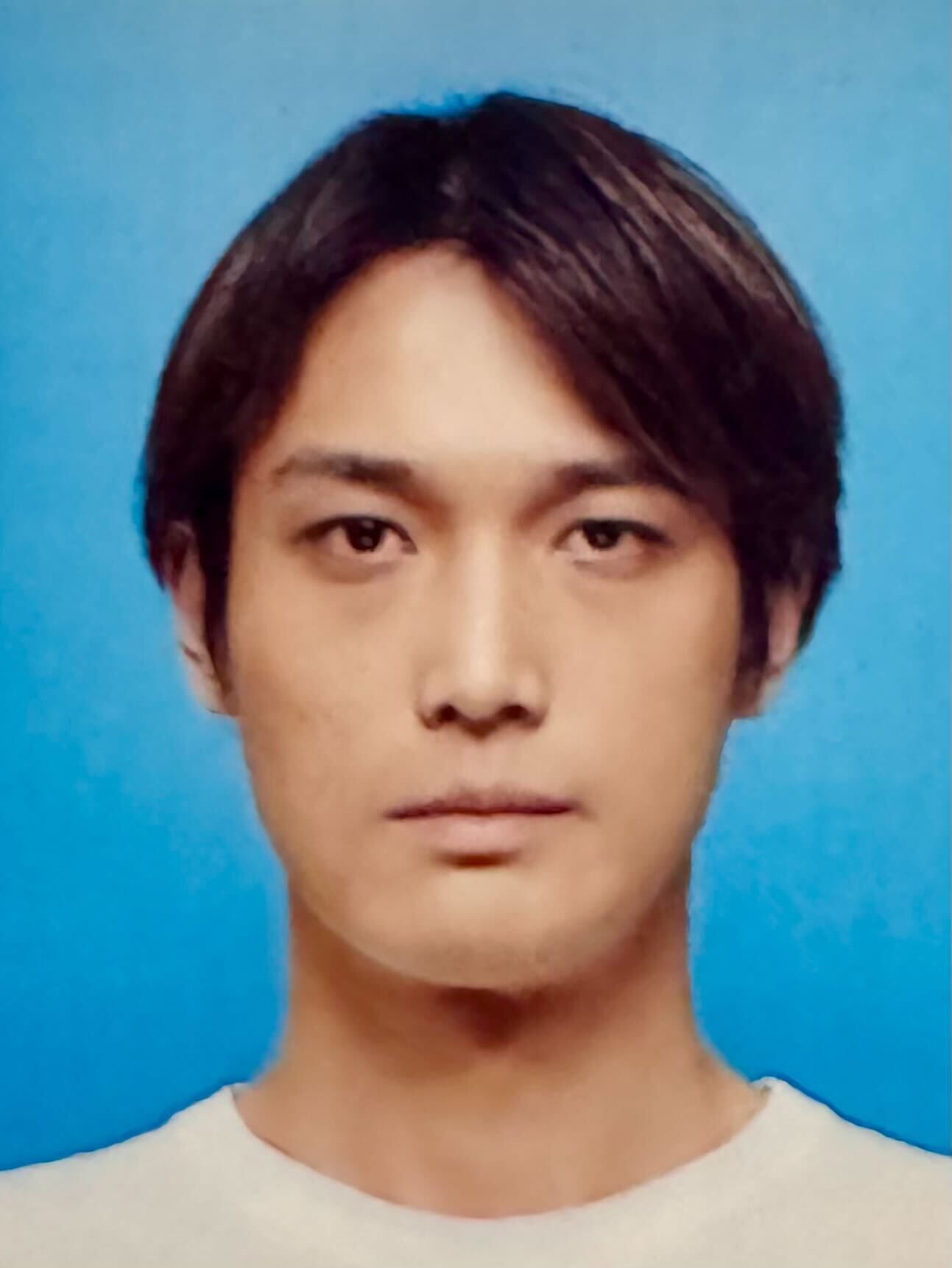}}]{Xiaotian Zhang}
(Member, IEEE) is a Ph.D. student in the Department of Precision Engineering, School of Engineering, The University of Tokyo, Japan. He received his M.S. degree in Precision Engineering from The University of Tokyo in 2021.

His research interests include robot perception and computer vision.
\end{IEEEbiography}

\newpage
\makeatletter
\def\@IEEEBIOskipN{12pt}
\makeatother
\noindent\begin{minipage}[t][5.6in][s]{\columnwidth}
\vspace{-12pt plus -1fil}
\begin{IEEEbiography}[{\includegraphics[width=1in,height=1.25in,clip,keepaspectratio]{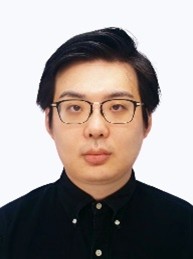}}]{Yusheng Wang}
(Member, IEEE) received the B.E. degree in mechanical engineering from the Dalian University of Technology, Dalian, China, in 2017, and the M.S. and Ph.D. degrees in precision engineering from The University of Tokyo, Bunkyo, Japan, in 2019 and 2022, respectively.

From 2022 to 2023, he was a Project Assistant Professor with the Department of Precision Engineering, The University of Tokyo. He is currently an Assistant Professor with Research into Artifacts, Center for Engineering, The University of Tokyo. His research interests include robot perception, computer vision, human-robot interaction, and marine robotics.

Dr. Wang is a Member of SICE, JSME, and RSJ.
\end{IEEEbiography}

\begin{IEEEbiography}[{\includegraphics[width=1in,height=1.25in,clip,keepaspectratio]{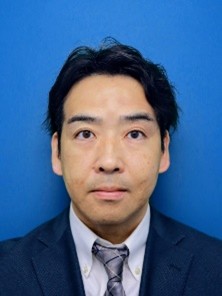}}]{Naoya Kagawa}
is a Manager in Factory Products Business Unit, Engineering Division, DENSO WAVE INCORPORATED, Japan.
\end{IEEEbiography}

\begin{IEEEbiography}[{\includegraphics[width=1in,height=1.25in,clip,keepaspectratio]{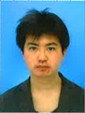}}]{Noritaka Takamura}
is an Engineer in Factory Products Business Unit, Engineering Division, DENSO WAVE INCORPORATED, Japan.
\end{IEEEbiography}

\end{minipage}
\newpage
\noindent\begin{minipage}[t][5.6in][s]{\columnwidth}
\vspace{-12pt plus -1fil}
\begin{IEEEbiography}[{\includegraphics[width=1in,height=1.25in,clip,keepaspectratio]{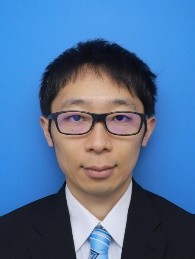}}]{Keiji Okuhara}
is an Engineer in Factory Products Business Unit, Engineering Division, DENSO WAVE INCORPORATED, Japan.
\end{IEEEbiography}

\begin{IEEEbiography}[{\includegraphics[width=1in,height=1.25in,clip,keepaspectratio]{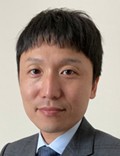}}]{Hiroyasu Baba}
is a Manager in Factory Products Business Unit, Strategy Planning Division, DENSO WAVE INCORPORATED, Japan.
\end{IEEEbiography}

\begin{IEEEbiography}[{\includegraphics[width=1in,height=1.25in,clip,keepaspectratio]{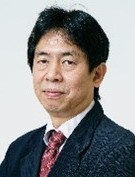}}]{Jun Ota}
(Member, IEEE) received the B.E., M.E., and Ph.D. degrees from the Faculty of Engineering, The University of Tokyo, Bunkyo, Japan, in 1987, 1989, and 1994, respectively.

From 1989 to 1991, he worked with Nippon Steel Corperation. In 1991, he was a Research Associate with The University of Tokyo. He became a Lecturer and Associate Professor in 1994 and 1996, respectively. In April 2009, he became a Professor with the Graduate School of Engineering, The University of Tokyo. In June 2009, he became a Professor with Research into Artifacts, Center for Engineering, (RACE), The University of Tokyo. From 1996 to 1997, he was a Visiting Scholar with Stanford University. From 2015 to 2021, he was a Guest Professor with the South China University of Technology. He is a Professor with RACE, School of Engineering, The University of Tokyo. His research interests include multiagent robotic systems, embodied-brain systems science, design support for large-scale production/material handling systems, and human behavior analysis and support.

Dr. Ota was the recipient of the fellowships from the Robotics Society of Japan in 2016 and from the Japan Society of Mechanical Engineers in 2021.
\end{IEEEbiography}

\end{minipage}
\end{document}